%% file: main.tex
\documentclass[10pt,twocolumn,letterpaper]{article}

\usepackage[pagenumbers]{cvpr} 

\input{preamble}

\definecolor{cvprblue}{rgb}{0.21,0.49,0.74}
\usepackage[pagebackref,breaklinks,colorlinks,citecolor=cvprblue,urlcolor=blue]{hyperref}

\def\paperID{269} 
\def\confName{3DV\xspace}
\def\confYear{2027\xspace}

\title{\sysname{}: 
A Large-Scale Real Cloth Material Dataset \\
for Neural Reflectance Reconstruction}

\author{%
\begin{tabular}[t]{@{}c@{}}
Zhen (Colin) Li$^{1*}$ \quad
Haoran Qin$^{1,2*\dagger}$ \quad
Francesco Di Sario$^{3,1}$ \quad
Daniel Rebain$^{4,5}$ \quad
Andrea Tagliasacchi$^{1,5,6}$\\[4pt]
{\normalsize $^1$Simon Fraser University \quad $^2$Westlake University \quad $^3$University of Torino}\\
{\normalsize $^4$University of British Columbia \quad $^5$Wayve Labs \quad $^6$University of Toronto}\\[6pt]
{\normalsize\href{https://theialab.github.io/robocloth/}{\textcolor{blue}{\underline{\smash{https://theialab.github.io/robocloth}}}}}
\end{tabular}%
}

\begin{document}

\maketitle
{\renewcommand{\thefootnote}{\fnsymbol{footnote}}\footnotetext[1]{Equal contribution.}\footnotetext[2]{Work done during an internship at Simon Fraser University.}}
\input{sec/0_abstract}
\input{sec/1_intro}

\input{sec/2_related}
\input{sec/3_dataset}
\input{sec/4_benchmark}
\input{sec/5_experiments}
\input{sec/6_conclusion}
{
    \small
    \bibliographystyle{ieeenat_fullname}
    \bibliography{main}
}

\input{sec/7_supp}

\end{document}

%% file: preamble.tex
\usepackage[dvipsnames]{xcolor}

\definecolor{revorange}{rgb}{0.80,0.38,0.0}

\AtEndPreamble{\crefname{figure}{Fig.}{Figs.}}
\usepackage{paralist}

\usepackage{soul}
\setuldepth{foobar}

\usepackage{makecell}
\usepackage{microtype}              
\usepackage{fontawesome5}           
\usepackage{pifont}                 
\usepackage{tikz}\usetikzlibrary{calc}  
\newcommand{\cmark}{\ding{51}}      
\newcommand{\xmark}{\ding{55}}
\newcommand{\sysname}{RoboCloth}            

\definecolor{linkgh}{HTML}{24292F}
\definecolor{linkpg}{HTML}{2563EB}
\definecolor{linkds}{HTML}{F59E0B}

\makeatletter
\apptocmd{\@maketitle}{%
  \vspace{-2em}%
  \begin{center}%
    \includegraphics[width=\linewidth]{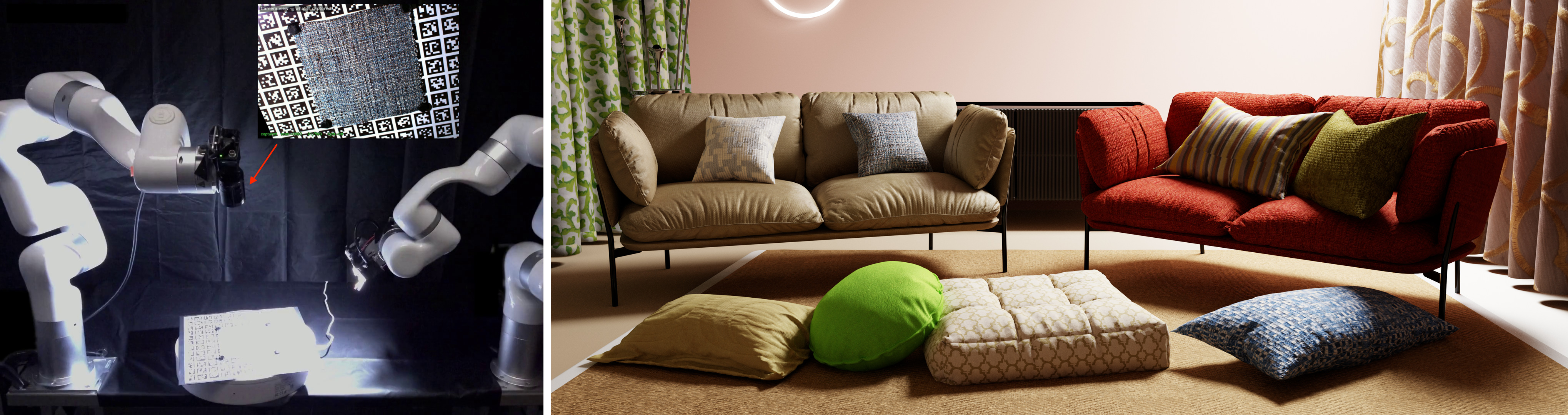}%
    \captionof{figure}{
    We design a robotic system to capture a large dataset of fabric materials, so to learn a neural BRDF representation.
    \emph{Left}: our capture system, two UFACTORY X-ARM 6 arms carrying a camera and a light source, with an example captured image shown in the inset.
    The photo of the rig is over-exposed so to be able to see the robotic arms, so the light saturates; captures use short exposures within the sensor's linear range.
    \emph{Right}: materials fitted via our learned prior on the test-set, yielding neural BRDFs that can be photo-realistically rendered with path-tracing~(i.e. the BRDFs of the fabric on the sofa, curtains, pillows, and carpet).}
    \label{fig:teaser}%
  \end{center}%
}{}{}
\def\paragraph{\@startsection{paragraph}{4}{\z@}%
  {1ex \@plus .3ex \@minus .2ex}{-1em}{\normalfont\normalsize\bfseries}}
\makeatother

%% file: sec/0_abstract.tex
\begin{abstract}
\vspace{-1em}
We present \sysname{}, a dataset of 500 real cloth materials captured under flexible, calibrated camera and lighting configurations, where both are robotically controlled during acquisition.
Each material is imaged in approximately 580 high-dynamic range frames spanning a wide view-light hemisphere; camera poses are refined via structure-from-motion and aligned to the robot frame, yielding per-material sparse surface geometry and structured per-point reflectance observations ready for neural material training.
We demonstrate the value of this data through a two-stage neural reconstruction pipeline:
Stage~1 trains a shared MLP decoder on our corpus to learn cloth material priors;
Stage~2 then freezes the decoder and fits a dense $2048{\times}2048$ latent texture to each target material.
Evaluating on held-out materials from our own dataset and on the independent UBO2014 benchmark, we show that a decoder trained on our captures outperforms decoders trained on existing measured reflectance datasets and analytic PBR fitting, transferring to unseen acquisition systems without modification.
The dataset, calibration files, observation tensors, and train/test splits are released under CC BY 4.0 on the Hugging Face Hub.
\end{abstract}

%% file: sec/1_intro.tex

\section{Introduction}
Cloth is challenging to render faithfully, as real fabrics exhibit spatially varying reflectance driven by fiber microstructure, weave geometry, surface roughness, and anisotropic scattering --- properties that interact nonlinearly with viewing and illumination direction and vary continuously across the surface.
Recovering dense, shader-ready material representations from real observations is a prerequisite for photorealistic cloth rendering in physically based pipelines, with direct applications in virtual try-on, digital fashion, and visual effects.
Recent work has shown that learning a \textit{shared} neural BRDF decoder across many materials enables generalizable reflectance estimation without committing to any analytical BRDF parameterization~\cite{rainer2020unified,raghavan2025generative}.
In this paradigm, a decoder maps compact per-point latent codes
to reflectance values queryable by a path tracer~\cite[Ch.~14]{pharr2016physically};
given a new material, a dense latent grid is optimized against sparse
observations with the decoder frozen, producing spatially varying
reflectance maps that plug directly into a physically based
rendering pipeline.
The decoder amortizes reflectance estimation across materials:
a prior trained on sufficiently diverse and well-covered captures
can generalize to unseen fabrics and recover dense representations
from limited measurements, a capability that neither single material overfitting nor purely analytical models can offer.

The limiting factor is training data.
Learning a generalizing decoder for cloth requires captures that
simultaneously satisfy three conditions:
\emph{real} measured appearance to avoid real-to-synth domain gaps,
\emph{diversity} in the captured materials to support generalization, and
\emph{dense+flexible} view-light coverage---directions sampled freely across
both hemispheres rather than from a preset fixed hardware layout---to enable
reliable reflectance-geometry disentanglement.
Existing resources do not combine real cloth diversity with flexible
view-light coverage at scale; see \Cref{tab:related}.
Most large-scale datasets of real materials are not multi-view: each surface is
observed from only one or two camera
positions~\cite{zhou2023photomat, ma2023opensvbrdf}.
Multi-view alternatives are either not spatially
varying~\cite{matusik2003data, jiri2014template}, or cover fewer than a
hundred materials~\cite{weinmann2014material}.
The closest resource to ours, UBOFAB19~\cite{merzbach2019learned},
provides 378 real fabrics with multi-view, spatially varying
measurements, but its view and light directions are dictated by a preset fixed
hardware layout rather than sampled freely across both hemispheres, which bounds the angular coverage of the view-light domain.
The cross-dataset results in \Cref{sec:results} show the performance drop caused by this insufficient view-light coverage.

To fulfill these requirements, we present \sysname{}, a system designed to capture a dataset of 500 materials.
Our system comprises two robot arms carrying an HDR camera and a calibrated
LED area light, together with a motorized turntable, yielding
$\approx 580$ HDR images per material; see \Cref{fig:teaser}.
View and light directions can be sampled freely across both hemispheres, rather than being constrained to a fixed set of hardware-defined configurations.
While approximate light and camera positions are controlled by the robot API, we recover accurate poses and sparse surface geometry via COLMAP~\cite{schonberger2016structure}.

Our dataset is not merely a larger image collection; it
directly targets a downstream task: recovering dense,
shader-ready material representations from real cloth captures
for a path-traced pipeline.
To this end, we train a neural BRDF decoder on our data and
evaluate it by fitting a dense per-pixel material map (a
latent texture) to each held-out material, measuring
reconstruction accuracy and appearance under
path-traced rendering, and comparing against classical PBR
fitting~\cite{burley2012physically} and decoders trained on existing material datasets.
A controlled ablation over dataset scale quantifies its
contribution to downstream generalization.

To summarize, our contributions are:
\begin{inparaenum}[(1)]
\item A real cloth reflectance dataset: 500 fabric materials with ${\approx}580$ HDR images each, under flexible view-light coverage, with pre-processed data, splits, and loaders for neural BRDF training; see \Cref{sec:dataset}.
\item A calibrated robotic acquisition pipeline: an end-to-end capture and processing system producing aligned HDR observations, camera and light metadata, COLMAP-refined poses, and sparse surface reconstructions; see \Cref{sec:dataset} and \Cref{supp:dataset}.
\item A neural decoder method for cloth material reconstruction:
training and evaluation of a neural BRDF decoder by
fitting a dense per-pixel latent texture to each held-out
material, with path-traced rendering evaluation and an ablation over dataset
scale; see \Cref{sec:experiments}.
\end{inparaenum}

\input{tables/related}

%% file: tables/related.tex
\begin{table}[t]
\centering
\renewcommand{\cmark}{\textcolor{ForestGreen}{\ding{51}}}%
\small
\setlength{\tabcolsep}{3pt}

\resizebox{\columnwidth}{!}{%
\begin{tabular}{lcrccccc}
\toprule
Dataset
& \makecell{Publication\\Year}
& \makecell{Dataset\\Scale}
& \makecell{Cloth\\Specific}
& \makecell{Real\\World}
& \makecell{Spatially\\Varying}
& \makecell{Multiple\\Views}
& \makecell{Flexible\\Coverage} \\
\midrule
MERL~\cite{matusik2003data}
& 2003 & 100 & \xmark & \cmark & \xmark & \cmark & \xmark \\

UTIA~\cite{jiri2014template}
& 2014 & 150 & \xmark & \cmark & \xmark & \cmark & \xmark \\

UBO2014~\cite{weinmann2014material}
& 2014 & 84 & \xmark & \cmark & \cmark & \cmark & \xmark \\

UBOFAB19~\cite{merzbach2019learned}
& 2019 & 378 & \cmark & \cmark & \cmark & \cmark & \xmark \\

Real-SVBRDF~\cite{asselin2020deep}
& 2020 & 80 & \xmark & \cmark & \cmark & \xmark & \xmark \\

PhotoMat~\cite{zhou2023photomat}
& 2023 & 12{,}000 & \xmark & \cmark & \cmark & \xmark & \xmark \\

UMat~\cite{rodriguez2023umat}
& 2023 & 2{,}000 & \cmark & \cmark & \cmark & \xmark & \xmark \\

OpenSVBRDF~\cite{ma2023opensvbrdf}
& 2023 & 1{,}000 & \xmark & \cmark & \cmark & \xmark & \xmark \\

MatSynth~\cite{vecchio2024matsynth}
& 2024 & 4{,}069 & \xmark & \xmark & \cmark & \xmark & \xmark \\

\midrule
\textbf{RoboCloth}
& -- & \textbf{500} & \cmark & \cmark & \cmark & \cmark & \cmark \\
\bottomrule
\end{tabular}%
}

\caption{
Comparison wrt.\ related works.
We report whether each dataset is cloth-specific and provides real measurements,
spatially varying appearance, multiple views, and flexible view-light coverage.
With multiple views, we denote capture with more than two distinct viewing directions.
RoboCloth is the only dataset to provide all five, and at scale.
}
\label{tab:related}
\end{table}

%% file: sec/2_related.tex
\section{Related Work}
We group prior work into four categories: measured reflectance and material datasets, inverse-rendering and relighting datasets, neural material representations, and calibrated capture systems.

\paragraph{Measured reflectance and material datasets.}
Classical BRDF databases such as MERL~\cite{matusik2003data} (100 isotropic
materials) provide dense per-sample reflectance but no spatial variation
across the surface. Denser angular collections are either not spatially varying or fixed to specialized hardware: the UTIA anisotropic BRDF
database~\cite{jiri2014template} (150 materials, 96 fabrics) stores one
spatially-averaged BRDF per material rather than a per-point SVBRDF.
Although its acquisition setup uses actuated camera and light arms, their
motion is constrained to fixed-radius goniometric trajectories rather than
arbitrary Cartesian poses, and all materials share the same uniform
$81{\times}81$ directional grid.
The UBO2014 BTF database~\cite{weinmann2014material} (84 materials,
22{,}801 images each) captures per-point appearance only under a fixed dome.
Among cloth-specific resources, UBOFAB19~\cite{merzbach2019learned} provides
378 fabric materials acquired with a commercial X-Rite TAC7 scanner under
a preset fixed hardware layout, providing at most 768
calibrated HDR measurements per material and Ward anisotropic SVBRDF fits
in AxF format.
Tang et al.~\cite{tang2024woven} capture woven-fabric BSDFs from a
reflection-transmission photo pair.
UMat~\cite{rodriguez2023umat} provides 2{,}000 real textiles for single-image
high-resolution material capture, while OpenSVBRDF~\cite{ma2023opensvbrdf}
scales to 1{,}000 near-planar SVBRDFs using a 16{,}384-LED rig and only two
camera views~(${\approx}193$ raw HDR images, ${\approx}15$~min per sample),
but is not cloth-specific.
Real-SVBRDF~\cite{asselin2020deep} covers 80 real materials across
462 capture sets with only 12 images per set.
Across these measured resources, none targets a large cloth corpus with
flexible view-light coverage (see \citet{kavoosighafi2024deep} for a broader
survey).
Moreover, none of these works compares against other datasets by
evaluating how well a prior trained on its data generalizes to new
materials.
On the synthetic side, MatSynth~\cite{vecchio2024matsynth} (4{,}069 PBR
materials) offers strong pretraining but does not replace real, calibrated cloth captures.

\paragraph{Inverse rendering and relighting datasets.}
Several datasets provide real object-level multi-illumination benchmarks useful for evaluating inverse rendering and relighting~\cite{kuang2023stanford, liu2023openillumination, ummenhofer2024objects, pei2025opensubstance, zhou2026olatverse}.
But these target \textit{object-centric} capture, and are neither designed for learning cloth material priors, nor for evaluating dense shader-ready material reconstruction from near-planar samples.

\paragraph{Neural material representations.}
NeuMIP~\cite{kuznetsov2021neumip} represents materials as multiscale
neural feature textures and has become a standard backbone for learned
material models, with subsequent work extending neural reflectance
representations through hierarchical architectures, efficient neural
sampling, real-time appearance models, and spherically distributed
primitives~\cite{xue2024hierarchical,zeltner2024real,weinreich2024real,xu2023neusample,dou2024real}.
Unified neural BTF encoding~\cite{rainer2020unified}, building on neural BTF
compression~\cite{rainer2019neural}, trains a decoder shared
across materials and supports encoding previously unseen materials.
DeepBRDF~\cite{hu2020deepbrdf} learns a latent space of measured BRDFs, and neural
BRDF representations with importance sampling~\cite{sztrajman2021neural}
make learned reflectance directly usable in path tracers.
HyperBRDF~\cite{gokbudak2024hypernetworks} and differentiable rendering
pipelines~\cite{deschaintre2018single,li2018materials} show that
reconstruction quality is tightly coupled to training data diversity
and angular coverage.
PhotoMat~\cite{zhou2023photomat} trains a strong generator from 12{,}000
single-flash real photos, underscoring the value of real captured data.
Generative Neural Materials~\cite{raghavan2025generative},
diffusion-based generators, and single-image capture
methods~\cite{vecchio2024matfuse, vecchio2024controlmat, rodriguez2023umat, zhang2024fabricdiffusion} all depend on large, diverse, and realistic training corpora.
Despite this progress, real captured datasets with cloth-specific
diversity and flexible view-light coverage remain limited.

\paragraph{Calibrated material capture systems.}
Capture systems trade off hardware, coverage, and throughput: dense
domes~\cite{weinmann2014material} and LED arrays~\cite{ma2023opensvbrdf}
give rich angular sampling but need dedicated, non-portable hardware.
Robot-based gonio-reflectometers such as ARGon3~\cite{hope2012argon3}
automate dense angular measurements of individually mounted samples.
Closest to our design is the BTF scanner of Rainer et al.~\cite{Rainer2019Highresolution},
which moves a camera and an LED light on two arms around a static sample and
acquires a dense $40{\times}40$ view-light scan in about 11 hours per sample;
its measurements were not publicly released.
Commercial scanners~\cite{merzbach2019learned} provide high-quality
fabric measurements under a preset fixed hardware layout.
Lightweight rigs~\cite{asselin2020deep} reduce hardware cost but yield
sparse per-sample observations.
Our robotic system---two robot arms carrying one HDR camera and one
calibrated LED area light, together with a motorized turntable---enables
practical cloth acquisition with hundreds of calibrated images per sample
under flexible view-light variation, without commitment to a fixed hardware
geometry or analytical BRDF model.

%% file: sec/3_dataset.tex
\section{Dataset}
\label{sec:dataset}
\sysname{} is a dataset of 500 real cloth materials, each captured under
approximately 580 calibrated camera-light configurations using a
robotic acquisition rig. For each material we release 16-bit linear HDR
images, robot-frame-aligned sparse geometry, a structured per-point
observation tensor, and the calibration metadata required to reproduce
our rendering pipeline. The dataset is split into training (${\approx}400$
materials) and test (${\approx}100$) partitions and is hosted on the Hugging Face Hub under \texttt{CC BY 4.0}.
Full construction details and ablations are provided in~\Cref{supp:dataset}.

\subsection{Acquisition}
\label{sec:acquisition}
As illustrated in \Cref{fig:teaser}, our robotic rig consists of two UFACTORY XARM~6 robot arms (one carrying the camera, one carrying an LED light source) and a ferromagnetic turntable inside a matte-black enclosure.
The turntable provides a third controlled degree of freedom: rotating the sample re-orients it beneath the fixed camera-light pair, so each arm configuration is reused across several distinct sample-relative azimuths, effectively tripling the diversity of camera-light configurations reachable without moving the arms.
Each cloth sample is fixed flat by corner magnets (standardizing the flattening procedure across operators), and a ChArUco fiducial~\cite{garrido2014automatic,romero2018speeded} mounted beside the sample serves for both offline calibration and per-session drift checks.
The sample width and length are recorded at mounting time and later used to crop the valid material region from the sparse point cloud.
Each material is acquired in a single automated session lasting approximately 20 minutes.
At each of the~${\approx} 580$ steps, the system records one raw Bayer HDR frame together with the robot-logged camera pose, light pose, and turntable angle, giving a densely paired image-and-metadata stream with no manual annotation.
Each pixel then contributes a per-point radiance observation:
In \Cref{sec:geometry} we describe how to place surface points, cameras, and lights in a common metric world frame, and in  \Cref{sec:radiometric} we calibrate the observed radiance to physical units.

\subsection{Geometry calibration and reconstruction}
\label{sec:geometry}
To place every surface point, camera, and light in a single metric
frame, a one-time hand-eye calibration~\cite{tsai1989new} recovers the camera-to-flange transform ${}^{g}T_c$, and a separate procedure fits the turntable rotation axis in the robot base frame from checkerboard-target images.
Composing these with the robot-logged arm pose gives each frame's camera pose in a common metric $0^\circ$-turntable world frame:
\[
{}^{w_0}T_{c,i}
= {}^{w_0}T_b(\theta_i)\,{}^{b}T_{g,i}\,{}^{g}T_c,
\]
where ${}^{w_0}T_b(\theta_i)$ is the Rodrigues rotation that undoes
the turntable angle $\theta_i$ about the calibrated axis, and the arm
pose ${}^{b}T_{g,i}$ is read directly from the robot controller's forward kinematics (its API) rather than estimated.
To recover a sparse point cloud, we employ COLMAP~\cite{schonberger2016structure} on low-dynamic-range~(LDR) copies of the frames.
Because the COLMAP reconstruction is defined only up to a similarity, we align it to the metric world frame in closed form via Umeyama~\cite{umeyama1991least} using matched camera
centers as correspondences.
Frames whose alignment residual exceeds $16\,\mathrm{mm}$ are discarded, and the sparse cloud is cropped to the known sample footprint and filtered in depth by the Tukey IQR rule~\cite{tukey1977exploratory}.

\subsection{Radiometric processing}
\label{sec:radiometric}
To make the observed radiance physically meaningful, raw Bayer
mosaics are de-mosaiced into linear-light HDR RGB using~\citet{menon2007demosaicing}; no gamma
or tone mapping is applied, preserving the raw radiance values recorded by the sensor.
We additionally calibrate a per-exposure camera factor $\kappa$ (a single scalar that converts pixel values to physical radiance under our linear sensor model) and the LED directional profile $\Lambda(\theta)$\footnote{A 90-entry polar table measured from a near-Lambertian reference card, which will be later used as $L_e^{\Lambda}$ in \cref{eq:rendering}}.
A $3{\times}3$ color correction matrix is estimated from a ColorChecker capture and reported alongside the release; the distributed images ship as white-balanced sensor RGB without the matrix applied, so downstream users retain full colorimetric control.

\subsection{Observation tensor}
\label{sec:obs_tensor}

Our training representation bypasses per-frame image loading entirely.
We project each retained sparse point into every retained HDR frame and bilinearly sample the corresponding RGB value, assembling a dense
tensor of \emph{surface points} $\mathbf{O}_m \in \mathbb{R}^{K_m \times V_m \times 3}$, where $K_m$ is the number of registered frames and $V_m$ the number of points.
Zero entries encode missing observations; no separate
visibility mask is needed. After filtering, each material yields
on average ${\approx} 1.6{\times}10^8$ valid RGB observations—a compact, memory-resident representation that replaces the ${\approx} 6$--$7\,\mathrm{GB}$ of per-material HDR frames (and their repeated per-frame decoding) with a single ${\approx} 1.0\,\mathrm{GB}$ indexed tensor of measurements.

%% file: sec/4_benchmark.tex
\section{Neural Reconstruction Pipeline}
\label{sec:neural}

We evaluate our dataset through a two-stage neural reconstruction pipeline, as illustrated in~\Cref{fig:stage1,fig:stage2}.
Stage~1 trains a shared MLP decoder on the full training
corpus to learn a material prior.
Stage~2 freezes that decoder and fits a dense per-material latent texture to each unseen cloth.
This freeze-and-fit scheme is similar to Generative Neural
Materials~\cite{raghavan2025generative}, which trains a shared decoder on
synthetic materials and fits per-material latent textures against the frozen
decoder to build training data for generative modeling.
The decoder architecture, renderer, and the Stage~2 training step~(losses, optimizer, and training schedule) are held fixed across all comparisons.

\input{figures/stage1}

\subsection{Differentiable Renderer}
\label{sec:renderer}

Both stages share the same differentiable single-bounce~\cite{kajiya1986rendering} path tracer, adopting the PyTorch differentiable path-tracer implementation from FIPT~\cite{wu2023factorized}.
Indirect illumination is omitted, as the sample sits in a matte-black enclosure with no secondary radiant sources within the
camera's dynamic range.
For a surface point $\mathbf{x}$ with outward normal $\mathbf{n}$,
the outgoing radiance toward the camera is
\begin{multline}
L_o(\mathbf{x}, \boldsymbol{\omega}_o) = \int_{\mathcal{E}}
f_r(\mathbf{x}, \boldsymbol{\omega}_i, \boldsymbol{\omega}_o)
\cdot L_e^{\Lambda}(\mathbf{y}, -\boldsymbol{\omega}_i)\,
\cdot \\
V(\mathbf{x},\mathbf{y})\,
\frac{(\boldsymbol{\omega}_i \cdot \mathbf{n})}{\|\mathbf{y}-\mathbf{x}\|^2}
\,\mathrm{d}A(\mathbf{y}),
\label{eq:rendering}
\end{multline}
where $L_e^{\Lambda}(\mathbf{y},-\boldsymbol{\omega}_i){=}L_e\,\Lambda(\theta)$ is the constant LED radiance shaped by the calibrated directional falloff $\Lambda(\theta)$~(\Cref{sec:radiometric}), and the
remaining terms account for geometry and visibility.
Because $\Lambda(\theta)$ already encodes the emitter's angular falloff, no separate light-side cosine appears in the integrand.
The finite-area integral is approximated with four Monte Carlo samples per surface point~($S{=}4$), which suffice given the LED's small size relative to the working distance.
The key design choice is that the neural decoder predicts \emph{only}
$f_r$, while cosine and geometry factors, emitter radiance, and inverse-square fall-off are all computed analytically from calibrated quantities.
This separation keeps the learned representation physically grounded and makes the frozen decoder directly portable to standard physically based renderers without \ul{any} modifications.

\subsection{Stage~1: Learning a Material Prior}
\label{sec:stage1}

Stage~1 trains a single MLP decoder $D_\theta$ that, conditioned on a
per-point latent code and local directions, predicts the BRDF value at
any queried configuration.
We adopt an auto-decoder formulation: each
sparse surface point $j$ of each training material $m$ owns three
directly optimizable quantities stored in aligned per-point parameter
tables---a latent
code $\mathbf{z}_{m,j} \in \mathbb{R}^{24}$, a surface normal
$\mathbf{n}_{m,j} \in \mathbb{R}^{3}$, and a tangent vector
$\mathbf{t}_{m,j} \in \mathbb{R}^{3}$ — initialized to $(0,0,1)$ and
$(0,1,0)$ respectively, consistent with the flat mounting geometry.
Before the decoder, a shared \emph{coordinate-transform operator}
$\mathcal{T}$ (\Cref{fig:stage1}) maps each point's
geometry and directions into the decoder's input: it forms a per-point
tangent-bitangent-normal (TBN) frame from $(\mathbf{n},\mathbf{t})$ by Gram--Schmidt, rotates the incident and outgoing directions
$\boldsymbol{\omega}_i,\boldsymbol{\omega}_o$ into that frame, encodes
the local directions and normal with degree-$3$ real spherical harmonics,
and concatenates those encodings with the latent code $\mathbf{z}$. The same $\mathcal{T}$
is used in both stages; its exact construction is given in~\Cref{supp:modules} and \Cref{fig:module_transform}.
The decoder $D_\theta$ is shared across all materials and all points;
the per-point parameter tables $(\mathbf{z}, \mathbf{n}, \mathbf{t})$ specialize
it to each surface location and are updated by the same optimizer step
as $\theta$. Training in this stage is \emph{sparse} and \emph{multi-material}:
each iteration draws a batch of (material, point, view) triplets sparsely
sampled across many different training materials, so the single shared
decoder is exposed to the full diversity of the corpus.
We evaluate \cref{eq:rendering} at each sampled observation using the
corresponding TBN frame, and back-propagate into both $\theta$ and the per-point parameter-table entries.
We learn our priors by optimizing the training objective:
$
\mathcal{L}
=
\mathcal{L}_{\text{recon}}
+
\lambda_{\text{decay}}\,\mathcal{L}_{\text{decay}},
$
whose terms we will now detail.

\paragraph{Reconstruction loss.}
As HDR captures span four to five orders of magnitude in radiance, and a
standard $\ell_1$ loss would be dominated by specular highlights, we
use a log-relative reconstruction loss, following the log-domain
mapping introduced for BRDF learning~\cite{nielsen2015optimal}:
\begin{equation}
\mathcal{L}_{\text{recon}}
=
\left|
\log\!\left(
  \tfrac{\hat{\mathbf{I}} + \varepsilon}{\rho_{\text{ref}} + \varepsilon} + 1
\right)
-
\log\!\left(
  \tfrac{\mathbf{I}_{\mathrm{GT}} + \varepsilon}{\rho_{\text{ref}} + \varepsilon} + 1
\right)
\right|,
\label{eq:recon}
\end{equation}
where $\hat{\mathbf{I}}$ is the rendered image obtained by evaluating
\cref{eq:rendering} at the observed configurations, $\mathbf{I}_{\mathrm{GT}}$ is the captured HDR image, $\rho_{\text{ref}}$ a per-dataset reference scale (\Cref{supp:impl}), and~$\varepsilon$ a small constant for numerical stability; we write this \textit{log-relative} transform as $T_{\log}(\cdot)$.
This gives equal weight to dark diffuse regions and bright specular
highlights across the full dynamic range.

\paragraph{Decay loss.}
Since captured observations do not cover incidence angles near grazing
($\cos\theta_i \approx 0$), the decoder is unconstrained in that
region. Physically, the contribution $f_r\cos\theta_i$ must vanish as
$\theta_i \to 90^\circ$. At each iteration, for a shared azimuth, we sample two incident directions
$\boldsymbol{\omega}_i^{h}$ and $\boldsymbol{\omega}_i^{\ell}$ with incidence-angle
cosines $c_h$ and $c_\ell = \alpha\,c_h$ ($\alpha \in [0,1]$): the anchor $c_h$ is
drawn from $[c_{\min},\,c_{\max}]$, just inside the most grazing observed angle, and
$c_\ell \le c_h$ places the second sample further toward grazing ($90^\circ$), in the
under-constrained region.
Defining $q(\boldsymbol{\omega}_i) = f_r(\boldsymbol{\omega}_i,
\boldsymbol{\omega}_o)\,\cos\theta_i$, we penalize
$q(\boldsymbol{\omega}_i^{\ell})$ whenever it exceeds the linear
interpolation between a value of zero at grazing, and the anchor
$q(\boldsymbol{\omega}_i^{h})$:
\begin{equation}
\mathcal{L}_{\text{decay}}
=
\Bigl[
  T_{\log}\!\bigl( q^{\ell} \bigr)
  \;-\;
  T_{\log}\!\bigl( \alpha\,
               \overline{q^{h}} \bigr)
\Bigr]_{+}\!,
\label{eq:decay}
\end{equation}
where $T_{\log}$ is the same log-relative transform as in~\cref{eq:recon}, $\overline{(\cdot)}$ denotes stop-gradient, and
$[\cdot]_{+}$ is ReLU.
We ablate this loss against alternatives in~\Cref{sec:ablation_grazing}.
The stop-gradient treats the anchor contribution $q^{h}$ as a fixed target, so the loss only suppresses the more-grazing prediction~$q^{\ell}$.
This encourages $f_r\cos\theta_i$ to decay as a concave function towards zero as we approach grazing angles, yet without forcing the raw BRDF itself to vanish.

\subsection{Stage~2: Dense Material Reconstruction}
\label{sec:stage2}
\input{figures/stage2}

%
While Stage~1 operates on sparse triplets $(\mathbf{z}, \mathbf{n}, \mathbf{t})$ provided by COLMAP points of multiple materials, Stage~2 is \emph{dense} and operates on a \emph{single} material.
For one target cloth, in Stage 2 we freeze $D_\theta$ and densely optimize a $2048\!\times\!2048$ \ul{latent-geometry grid} $\mathbf{Z}$ covering the full surface, where each texel stores the triplets alike those in Stage~1---latent code, normal, a tangent.
Every retrieved triplet is passed through the same $\mathcal{T}$ and frozen decoder as in Stage~1.
A progressive decaying Gaussian blur is applied to $\mathbf{Z}$ during
optimization, acting as a coarse-to-fine regularizer that suppresses
high-frequency overfitting early in training without an explicit penalty term.
The detailed implementation is provided by
\texttt{LatentTexture.apply\_gaussian\_blur} in
\texttt{model/neural\_brdf\_refactored.py} of the released code.

\paragraph{Texture-space displacement.}
As discussed in NeuMIP~\cite{kuznetsov2021neumip}, under a
flat-surface assumption, when the mesostructure has non-negligible height the
point actually visible along a given direction is laterally displaced from the
ray-surface intersection coordinate.
Thus, similar to NeuMIP, Stage~2 includes a \emph{parallax-aware texture lookup} $P$ (\Cref{fig:stage2}): a small per-material
geometry MLP predicts a
view- and light-dependent texture-space displacement
$\Delta\mathbf{u}(\mathbf{u},\boldsymbol{\omega}_i,\boldsymbol{\omega}_o)$
from the geometry latent at $\mathbf{u}$ and warps the coordinate before
the point attributes are bilinearly sampled from $\mathbf{Z}$.
The geometry MLP and $\mathbf{Z}$ are optimized jointly while $D_\theta$
stays frozen; the exact lookup is given in \Cref{supp:modules} and \Cref{fig:module_lookup}.
The same renderer and log-relative loss from Stage~1 are used throughout.
For the sparse baseline datasets, which provide no shared surface parametrization, Stage~2 reverts to the per-point representation of
Stage~1 (\Cref{supp:training_sources}).

%% file: figures/stage1.tex
\begin{figure*}[t]
  \centering
  \includegraphics[width=\linewidth]{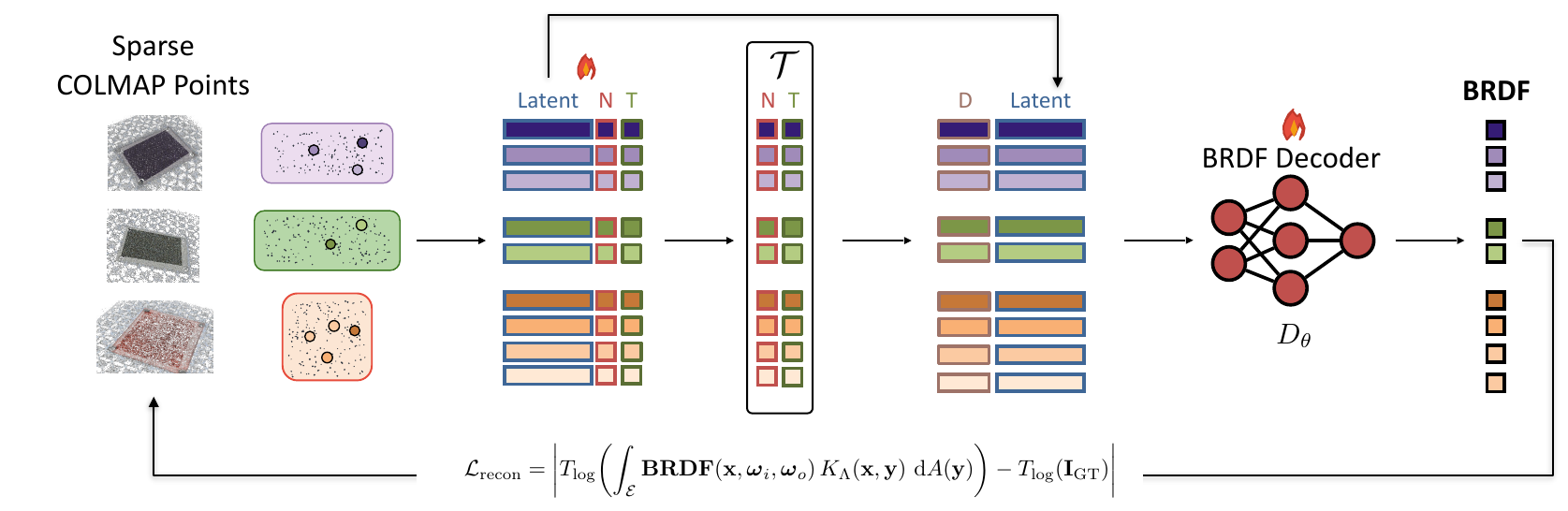}
  \caption{\textbf{Stage~1 --- learning a material prior.}
    A single BRDF decoder, shared across all training materials, is
    optimized jointly (flame) with per-point tables of latent codes,
    normals, and tangents; see \Cref{sec:stage1}.
    N: normal, T: tangent, D: encoded local directions; each color
    denotes one material, with a few points drawn per material in every
    minibatch.}
  \label{fig:stage1}
\end{figure*}

%% file: figures/stage2.tex
\begin{figure*}[t]
  \centering
  \includegraphics[width=\linewidth]{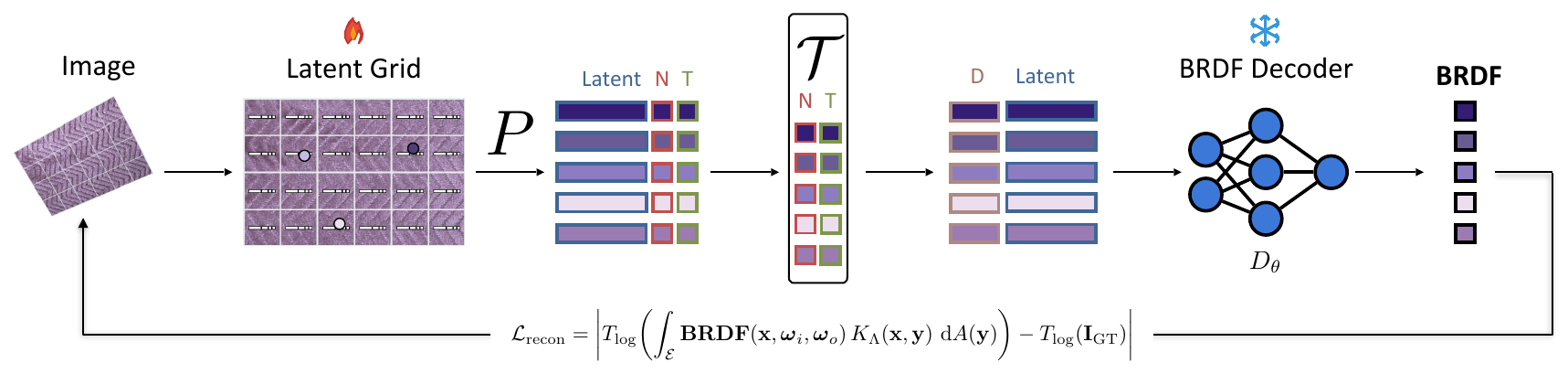}
  \caption{\textbf{Stage~2 --- dense material reconstruction.}
    For a single target material, a dense latent-geometry texture is
    optimized (flame) through the parallax-aware lookup $P$, the shared
    transform $\mathcal{T}$, and the frozen decoder (snowflake); see
    \Cref{sec:stage2}.
    Notation as in \Cref{fig:stage1}.
    }
  \label{fig:stage2}
\end{figure*}

%% file: sec/5_experiments.tex
\section{Experiments}
\label{sec:experiments}

To measure the value of different training corpora for Stage~1,
we use a controlled experimental protocol: the decoder architecture,
renderer, and Stage~2 procedure are held fixed across all comparisons,
so differences in reconstruction quality are \textit{directly} attributable to the training data.
We target two key questions:
$\quad$
\textit{Does training on our captured dataset produce a stronger material prior than training on existing reflectance datasets?}
$\quad$
\textit{Does the learned prior transfer to unseen cloth materials from an independent capture system?}
The complete implementation of both stages is publicly available at \url{https://github.com/theialab/robocloth}, so that every experiment reported below can be reproduced.

\subsection{Datasets and Baselines}
\label{sec:datasets}
%
We compare three Stage~1 training sources, all consumed through the same auto-decoder loop: \texttt{RoboCloth} (our real robotic captures), \texttt{Bonn}~\cite{merzbach2019learned} (the UBOFAB19 fabric SVBRDFs), and \texttt{MERL}~\cite{matusik2003data} (real, measured isotropic BRDFs). Per-source statistics and loading details are in~\Cref{supp:training_sources} and~\Cref{tab:training_sources}.

\paragraph{Baseline (PBR).}
As a non-neural reference we fit spatially-varying Disney principled
BRDF~\cite{burley2012physically} texture maps with the same renderer and optimizer
as Stage~2. The model sums an anisotropic GGX specular lobe, a Burley
diffuse lobe, and a sheen lobe; the optimized parameters are
$2048\!\times\!2048$ textures of base color, ambient occlusion,
roughness, metallic, IOR, anisotropy (strength and rotation), specular
tint, sheen, and sheen tint, plus a normal map, at the same resolution
as the neural latent grid. The full BRDF and its component definitions
are given in~\Cref{supp:pbr}.

\subsection{Implementation Details}
\label{sec:impl}
The decoder architecture, the direction encoding, the per-point and dense representations, the two-stage optimizers, and the full training and fitting schedules are reported in~\Cref{supp:impl}.
The complete code for both stages is publicly released, so that all reported experiments can be reproduced and fine-grained technical questions can be answered directly from the implementation.

\subsection{Results}
\label{sec:results}
All evaluations in this section are based on Stage-2 fitting results: the only metric available in Stage~1 is the training loss on the different training datasets, so the Stage-2 fitting is needed to evaluate the decoder on the test set.
We evaluate the learned material prior in three regimes, varying only the Stage~1 training source while holding the decoder, renderer, and Stage~2 fixed: \textit{cross-dataset transfer} to an independent capture system (UBO2014), \textit{in-domain reconstruction} on held-out materials from our own test set, and \textit{cross-dataset comparison} on held-out Bonn materials.
In every table the three decoder columns differ only in their Stage-1 training source (RoboCloth\,/\,Bonn\,/\,MERL); PBR is the analytic Disney principled fitting baseline. We report per-material PSNR$\uparrow$~(dB).
Fitting a dense latent grid at high resolution takes a long time (up to one day of training on an RTX 6000 Ada).
Thus, due to computation-resource limits, we randomly pick a subsample of materials for evaluation: 12 from UBO2014, 5 from our own held-out test set, and 5 from the Bonn test set.

\paragraph{Cross-dataset transfer to UBO2014 -- \Cref{tab:ubo}.}
We freeze each pretrained decoder and fit it to UBO2014. Because UBO2014 provides sparse per-point ray-camera observations rather than a continuously parametrized surface, we use the Stage-1-style per-point latent representation together with the per-channel scale
$\boldsymbol{\beta}$~{(\Cref{supp:impl})}, without the dense latent-geometry grid or parallax-aware texture lookup $P$~(\Cref{sec:stage2}).
We evaluate 12 materials randomly selected from the UBO2014 dataset~\cite{weinmann2014material}.
UBO2014 provides $151{\times}151 \approx 2.3{\times}10^4$ angular samples per spatial pixel at $400{\times}400$ resolution.
The decoder trained on our captures attains the highest PSNR on \emph{every one} of the 12 materials, leading (by a large margin) the next-best baseline (PBR) by $3.7$~dB and the strongest competing decoder (Bonn) by $4.5$~dB on average.
This is despite never seeing a UBO2014 sample during Stage~1, and it is direct evidence that the flexible, calibrated view-light coverage of our captures yields a more transferable cloth prior than existing measured (Bonn, MERL) or analytic (PBR) alternatives.

\input{tables/ubo}

\paragraph{Qualitative comparison.}
Relighting the held-out UBO2014 materials in a common scene (same mesh, UV tiling, lighting, and camera for every method, so any visible difference is due \textit{solely} to the learned material) shows the same ordering as the table:
the MERL-trained decoder, trained only on spatially-uniform isotropic BRDFs, loses the weave structure and drifts in color;
the Bonn-trained decoder recovers texture but shifts the overall hue, and the analytic PBR baseline misses the fine sheen;
only our decoder matches the ground-truth color while preserving both the specular highlight at the folds and the fuzzy weave detail, and it attains the highest image-space PSNR on every rendered view.
All qualitative results --- renders of all 12 UBO2014 materials, the per-test-set reconstructions, and image-space comparisons against the captured ground truth --- are provided on the project page (\url{https://theialab.github.io/robocloth/}).

\paragraph{In-domain SVBRDF -- \Cref{tab:ours}~(top).}
We next evaluate dense reconstruction on materials held out from our own dataset: for each material we freeze the decoder, optimize a dense latent grid on a subset of captured views, and evaluate on held-out views and lighting. As shown in \Cref{tab:ours}, top, the RoboCloth-trained decoder wins on every material, confirming that the prior is well matched to our capture distribution.

\input{tables/ours}

\paragraph{Cross-domain SVBRDF -- \Cref{tab:ours}~(bottom).}
Finally, we test transfer to the Bonn distribution.
As expected, the Bonn-trained decoder ranks first here, since these materials are held out from its own training distribution and the test set is therefore in-distribution for it.
Our decoder is a consistent \emph{second}, ahead of
both the MERL-trained decoder and the PBR baseline on every material:
even off-distribution, the prior learned from our captures transfers
more effectively than the other non-native models. The ranking
Bonn~${>}$~RoboCloth~${>}$~PBR~${>}$~MERL holds on all five materials.
Crucially, this transfer is \emph{asymmetric}: on Bonn's own
distribution our decoder trails the native model by only $1.0$~dB
($23.43$ vs.\ $24.47$), whereas on our distribution the Bonn-trained
decoder trails ours by $2.3$~dB ($27.32$ vs.\ $29.63$)---more than twice
the gap. A decoder trained on \sysname{} therefore comes close to matching a Bonn-native decoder on Bonn, while \textit{the converse does not hold}.

\paragraph{Ablations.}
We ablate the Stage-1 training-set size: retraining the
decoder on random $100/300/500$-material subsets ($80/240/400$ training materials), and re-running the UBO2014 transfer shows PSNR rising consistently with more data and beginning to saturate at our full corpus. Please refer to the supplementary \Cref{supp:datasize,fig:datasize,tab:datasize}.

%% file: tables/ubo.tex
\begin{table}[t]
\centering
\footnotesize
\renewcommand{\arraystretch}{0.92}
\setlength{\tabcolsep}{3.5pt}
\begin{tabular}{l cccc c}
\toprule
& \multicolumn{3}{c}{Decoder training source} & Baseline & \\
\cmidrule(lr){2-4}\cmidrule(lr){5-5}\cmidrule(lr){6-6}
Material & RoboCloth & Bonn & MERL & PBR & {vs.\ best} \\
\midrule
Fabric02 & \textbf{34.57} & 28.56 & 27.88 & \underline{30.96} & {$+3.61$} \\
Fabric04 & \textbf{34.00} & 28.50 & 26.66 & \underline{30.09} & {$+3.91$} \\
Fabric09 & \textbf{39.50} & 36.08 & 33.96 & \underline{36.96} & {$+2.54$} \\
Fabric11 & \textbf{34.33} & 29.45 & 27.46 & \underline{31.24} & {$+3.09$} \\
Felt01   & \textbf{38.76} & 33.60 & 32.81 & \underline{35.05} & {$+3.71$} \\
Felt03   & \textbf{40.30} & 34.20 & 33.27 & \underline{36.66} & {$+3.64$} \\
Felt05   & \textbf{35.39} & 32.13 & 30.32 & \underline{32.63} & {$+2.76$} \\
Felt10   & \textbf{44.36} & 38.52 & 36.38 & \underline{41.10} & {$+3.26$} \\
Carpet02 & \textbf{38.87} & \underline{34.82} & 32.56 & 33.75 & {$+4.05$} \\
Carpet07 & \textbf{34.61} & \underline{30.17} & 28.85 & 29.48 & {$+4.44$} \\
Carpet09 & \textbf{33.17} & \underline{31.23} & 29.42 & 29.97 & {$+1.94$} \\
Carpet12 & \textbf{33.25} & \underline{29.93} & 28.13 & 28.70 & {$+3.32$} \\
\midrule
\textbf{Average} & \textbf{36.76} & 32.27 & 30.64 & \underline{33.05} & {$+3.71$} \\
\bottomrule
\end{tabular}
\caption{\textbf{Cross-dataset transfer to UBO2014.}
Decoder columns differ only in Stage-1 training source;
PSNR in dB ($\uparrow$).
The last column reports the margin of RoboCloth over the best
competing column (positive means RoboCloth leads).
}
\label{tab:ubo}
\end{table}

%% file: tables/ours.tex
\begin{table}[t]
\centering
\footnotesize
\renewcommand{\arraystretch}{0.92}
\setlength{\tabcolsep}{3.5pt}
\begin{tabular}{l cccc c}
\toprule
& \multicolumn{3}{c}{Decoder training source} & Baseline & \\
\cmidrule(lr){2-4}\cmidrule(lr){5-5}\cmidrule(lr){6-6}
Material & RoboCloth & Bonn & MERL & PBR & {vs.\ best} \\
\midrule
\multicolumn{6}{l}{\textit{In-domain: \sysname{} test set (held-out)}} \\
\midrule
226 & \textbf{29.24} & \underline{25.63} & 24.24 & 25.35 & {$+3.61$} \\
314 & \textbf{26.60} & \underline{24.42} & 22.93 & 23.98 & {$+2.18$} \\
370 & \textbf{29.71} & 29.21 & 27.75 & \underline{29.29} & {$+0.42$} \\
145 & \textbf{28.44} & \underline{24.45} & 23.39 & 24.28 & {$+3.99$} \\
452 & \textbf{34.15} & \underline{32.91} & 30.89 & 32.06 & {$+1.24$} \\
\textbf{Average} & \textbf{29.63} & \underline{27.32} & 25.84 & 26.99 & {$+2.31$} \\
\midrule
\multicolumn{6}{l}{\textit{Cross-domain: Bonn test set (held-out)}} \\
\midrule
318 & \underline{44.38} & \textbf{45.83} & 40.26 & 43.16 & {$-1.45$} \\
377 & \underline{16.29} & \textbf{16.78} & 15.57 & 15.93 & {$-0.49$} \\
32  & \underline{15.31} & \textbf{16.42} & 14.18 & 15.17 & {$-1.11$} \\
226 & \underline{24.14} & \textbf{25.09} & 22.46 & 23.92 & {$-0.95$} \\
37  & \underline{17.03} & \textbf{18.22} & 16.36 & 16.75 & {$-1.19$} \\
\textbf{Average} & \underline{23.43} & \textbf{24.47} & 21.77 & 22.99 & {$-1.04$} \\
\bottomrule
\end{tabular}
\caption{\textbf{Per-material reconstruction.} \emph{Top:}
in-domain on our held-out test set. \emph{Bottom:} cross-dataset on
held-out Bonn materials.
The last column reports the margin of RoboCloth over the best
competing column, in dB; positive means RoboCloth leads.
}
\label{tab:ours}
\end{table}

%% file: sec/6_conclusion.tex
\section{Conclusion}

We presented \sysname{}, a dataset of 500 real cloth materials captured under flexible, calibrated camera and light coverage using a robotic acquisition system.
Our end-to-end pipeline produces per-material training records — aligned HDR images, sparse surface geometry, and structured observation tensors that feed directly into neural decoder training without manual annotation.
Under a controlled two-stage protocol in which only the Stage~1 training data varies, a decoder trained on our captures learns a stronger cloth prior than decoders trained on existing measured reflectance datasets, recovering more faithful held-out appearance and transferring competitively to the UBO2014 benchmark.

\paragraph{Limitations.}
The dataset is cloth-specific: the acquisition protocol and
evaluation are tailored to near-planar fabric samples and do not
generalize to three-dimensional objects or non-textile materials.
\quad
Acquisition throughput is modest: each material takes
${\approx} 20$ minutes and needs a human operator to mount and change the sample between sessions.
Thus, although in practice we have ${\approx}5000$ more cloth samples, throughput is currently limited by the most difficult resource to scale: the graduate student.
We are therefore considering a different capture modality for faster acquisition; throughput could also clearly be scaled by adding more robot arms.
Notably, the dataset-size ablation (\Cref{fig:datasize} in
\Cref{supp:datasize}) shows that transfer PSNR still improves at the
full corpus: learning has not saturated, and we stopped at $500$
materials simply as a round number, given the slow capture.
\quad
Since a material prior is available from previous scans, an
active-capture agent could select the next-best view-light configuration for each new sample, replacing the ${\approx}580$ uniformly scheduled frames with a small adaptive set and directly addressing the throughput bottleneck.
\quad
Finally, this work performs SVBRDF inversion for planar materials. How to integrate the resulting material priors effectively within a \textit{scene-level} 3D Gaussian Splatting reconstruction~\cite{kerbl20233d} remains an open research direction.




%% file: sec/7_supp.tex

\clearpage
\setcounter{page}{1}
\maketitlesupplementary


\section*{Supplementary Overview}

This supplementary material is organized as follows.
\Cref{sec:ablation_grazing} presents the grazing-angle
supervision ablation and \Cref{supp:datasize} the dataset-size
ablation. \Cref{supp:dataset} provides full details of the dataset
construction pipeline, including the capture trajectory design,
complete details of the hand-eye and turntable-axis
calibration, the Umeyama similarity alignment, and the radiometric
calibration procedures summarized in the main paper.
\Cref{supp:modules} expands the two pipeline operators---the
coordinate transform $\mathcal{T}$ and the parallax-aware texture lookup
$P$---that appear as labeled boxes in the main-paper pipeline
figures (Figs.~\ref{fig:stage1} and~\ref{fig:stage2}).
\Cref{supp:pbr} gives the full Disney principled BRDF used
by the PBR fitting baseline.
\Cref{supp:impl} reports the full set of implementation
hyperparameters for Stage~1 and Stage~2, together with additional
notes on the 8-bit Adam optimizer, the per-point optimizable
parameters, and the round-robin material sampling strategy.
\Cref{supp:dataset_schema} documents the schema of the
released Hugging Face dataset.
Additional qualitative results on held-out materials and UBO2014
transfers are provided on the project page
(\url{https://theialab.github.io/robocloth/}).

\section{Ablation: Grazing-Angle Supervision}
\label{sec:ablation_grazing}

Our captured observations do not cover near-grazing incidence
($\cos\theta_i\!\approx\!0$), leaving the decoder unconstrained there. The
grazing-decay loss of \cref{eq:decay} constrains the \emph{shape} of the
contribution $f_r\cos\theta_i$ to fall off smoothly to zero as
$\theta_i\!\to\!90^\circ$, without forcing the raw BRDF $f_r$ itself to
vanish. We compare it against two simpler pseudo-observation losses and a
no-grazing baseline. \textbf{ZeroExact} adds a pseudo-observation that pins
the raw BRDF to zero at \emph{exactly} the grazing angle
($\theta_i\!=\!90^\circ$). \textbf{NearZeroBRDF} pins the raw BRDF to zero
over a thin band just below grazing ($\cos\theta_i\!\in\![0.005,0.03]$).
\textbf{No grazing} applies no such supervision.

\begin{figure*}[t]
  \centering
  \setlength{\tabcolsep}{1.5pt}
  \renewcommand{\arraystretch}{0.6}
  \begin{tabular}{@{}cccc@{}}
  {\small Ours (Eq.~\ref{eq:decay})} & {\small ZeroExact} & {\small NearZeroBRDF} & {\small No grazing} \\
  \includegraphics[width=0.235\linewidth]{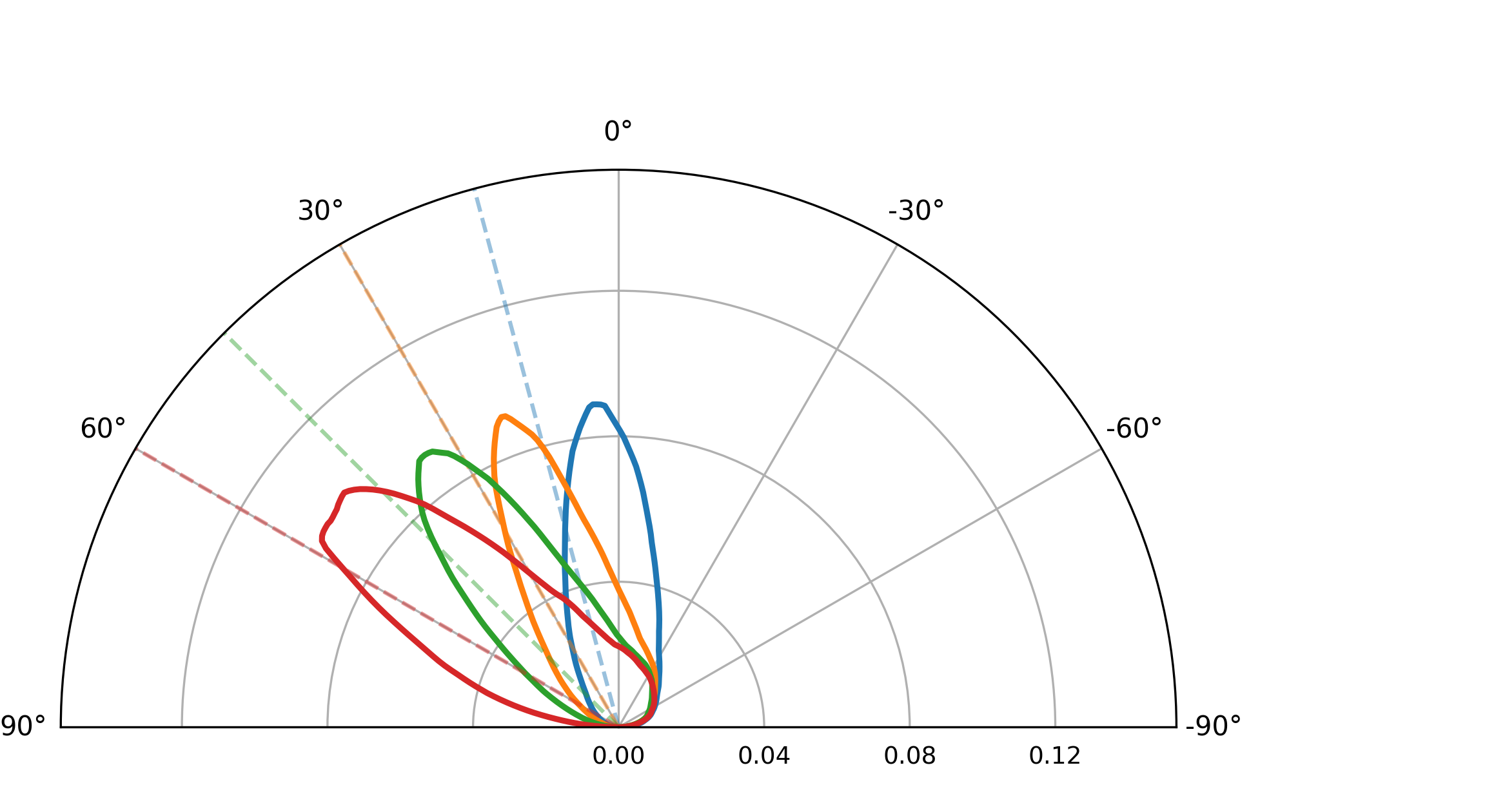} &
  \includegraphics[width=0.235\linewidth]{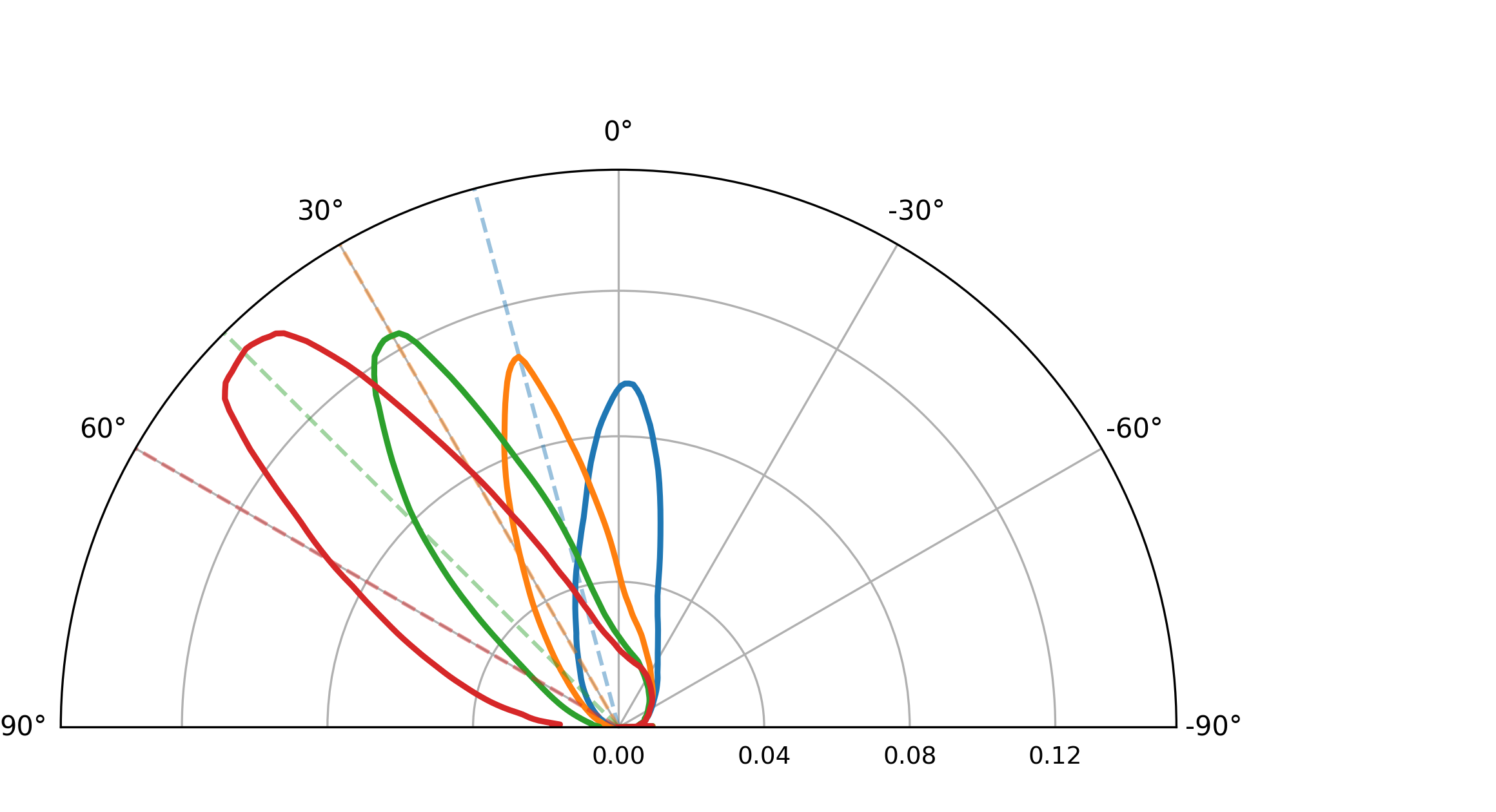} &
  \includegraphics[width=0.235\linewidth]{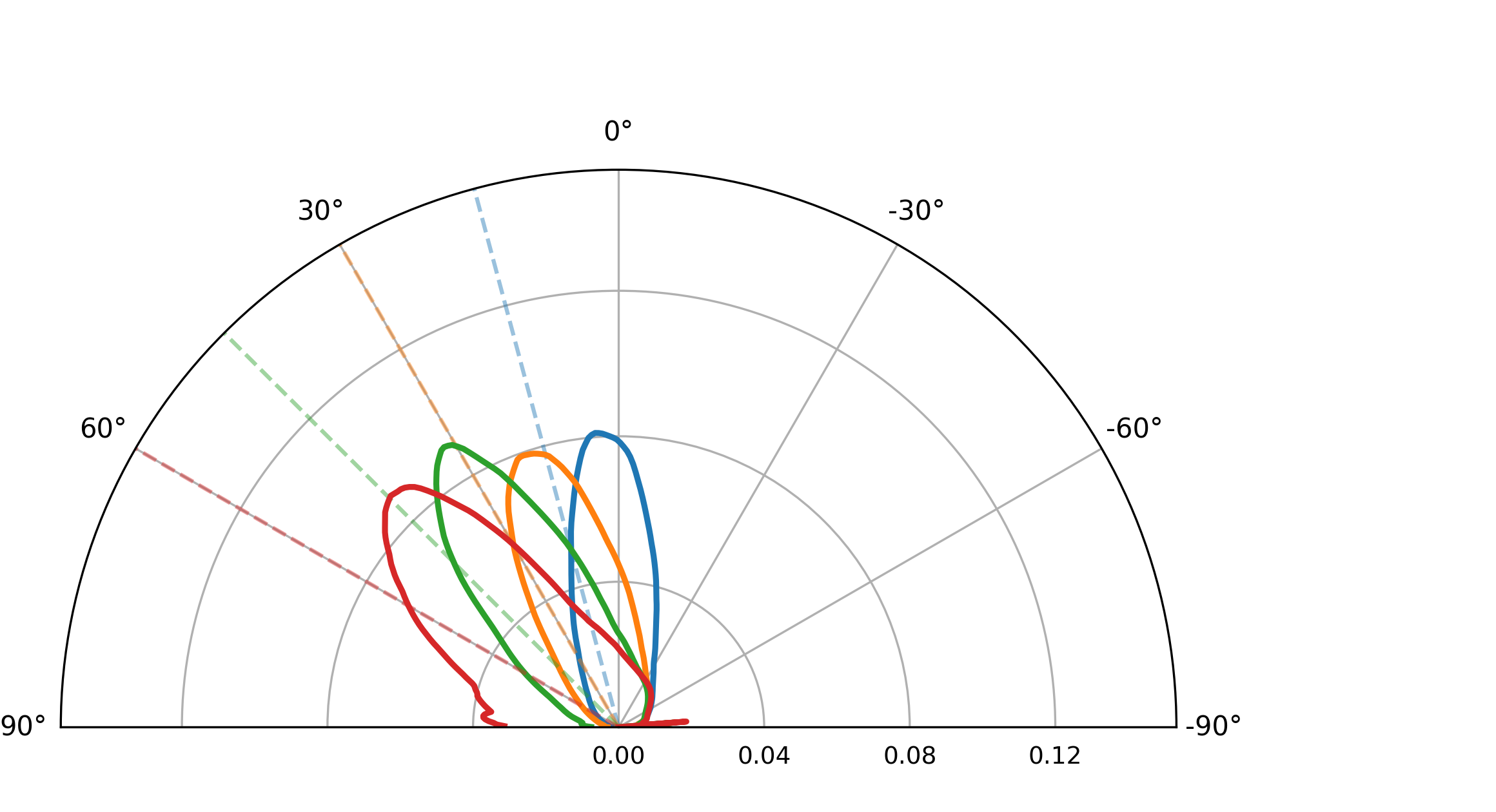} &
  \includegraphics[width=0.235\linewidth]{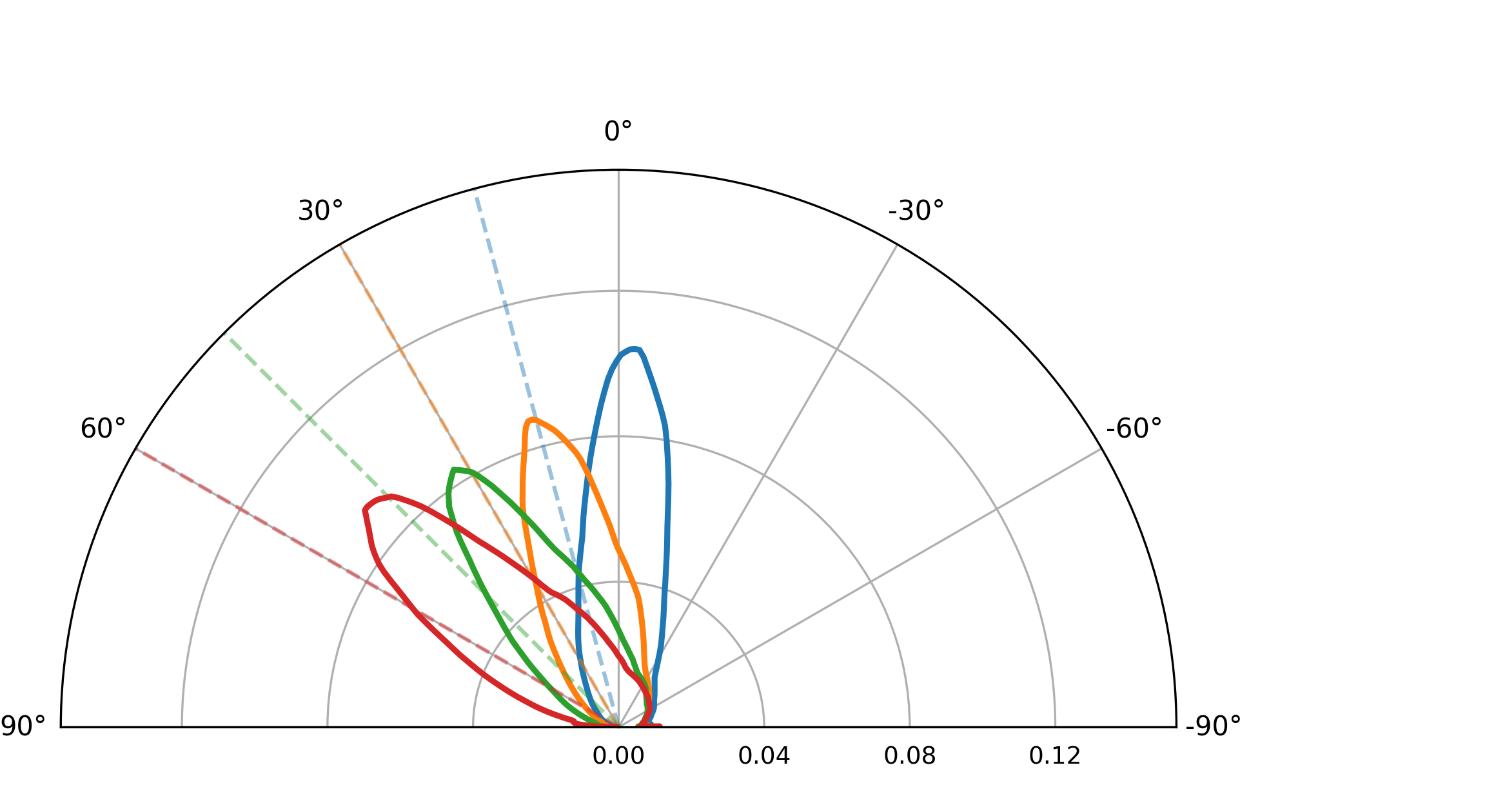} \\
  \includegraphics[width=0.235\linewidth]{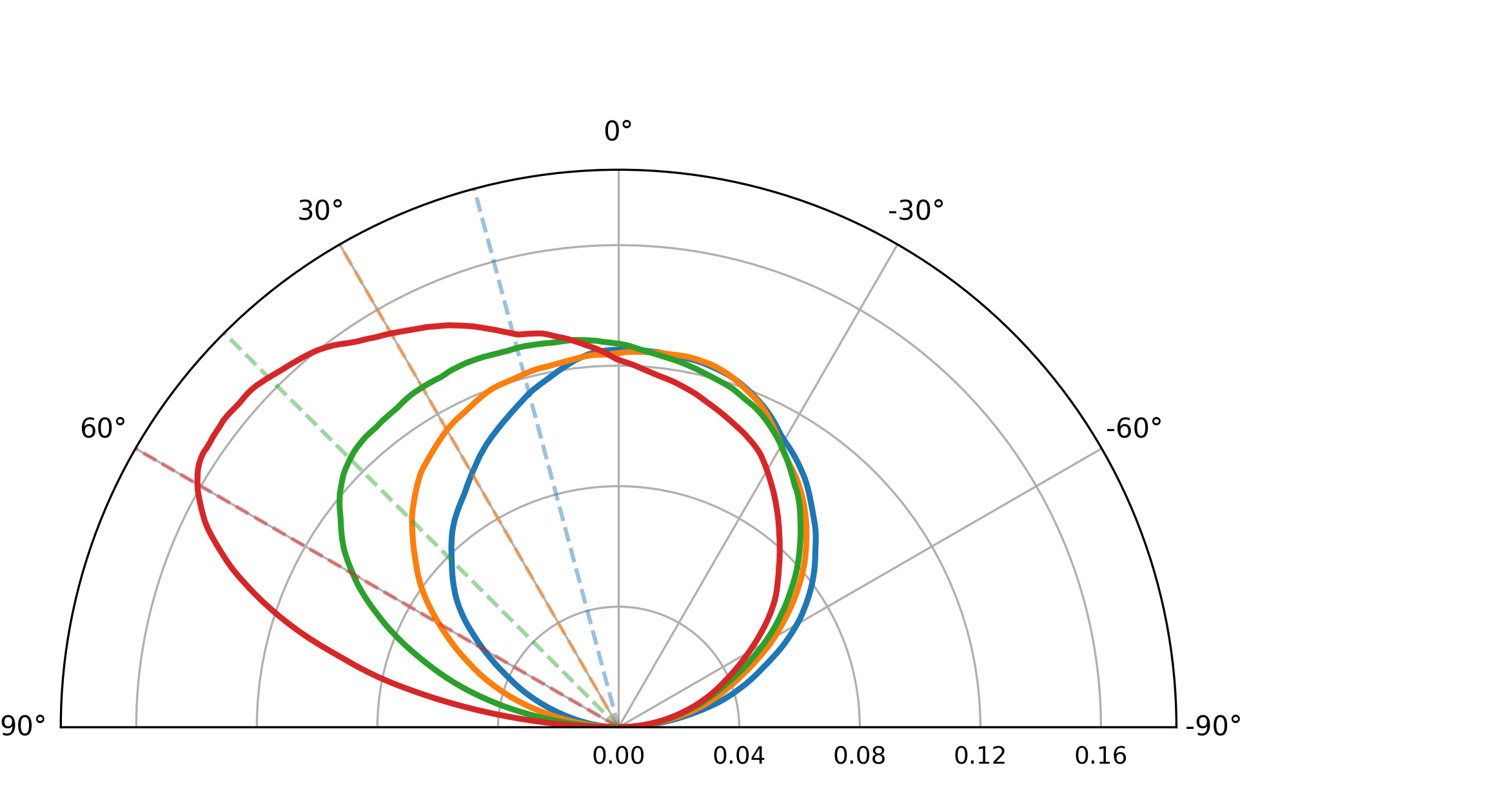} &
  \includegraphics[width=0.235\linewidth]{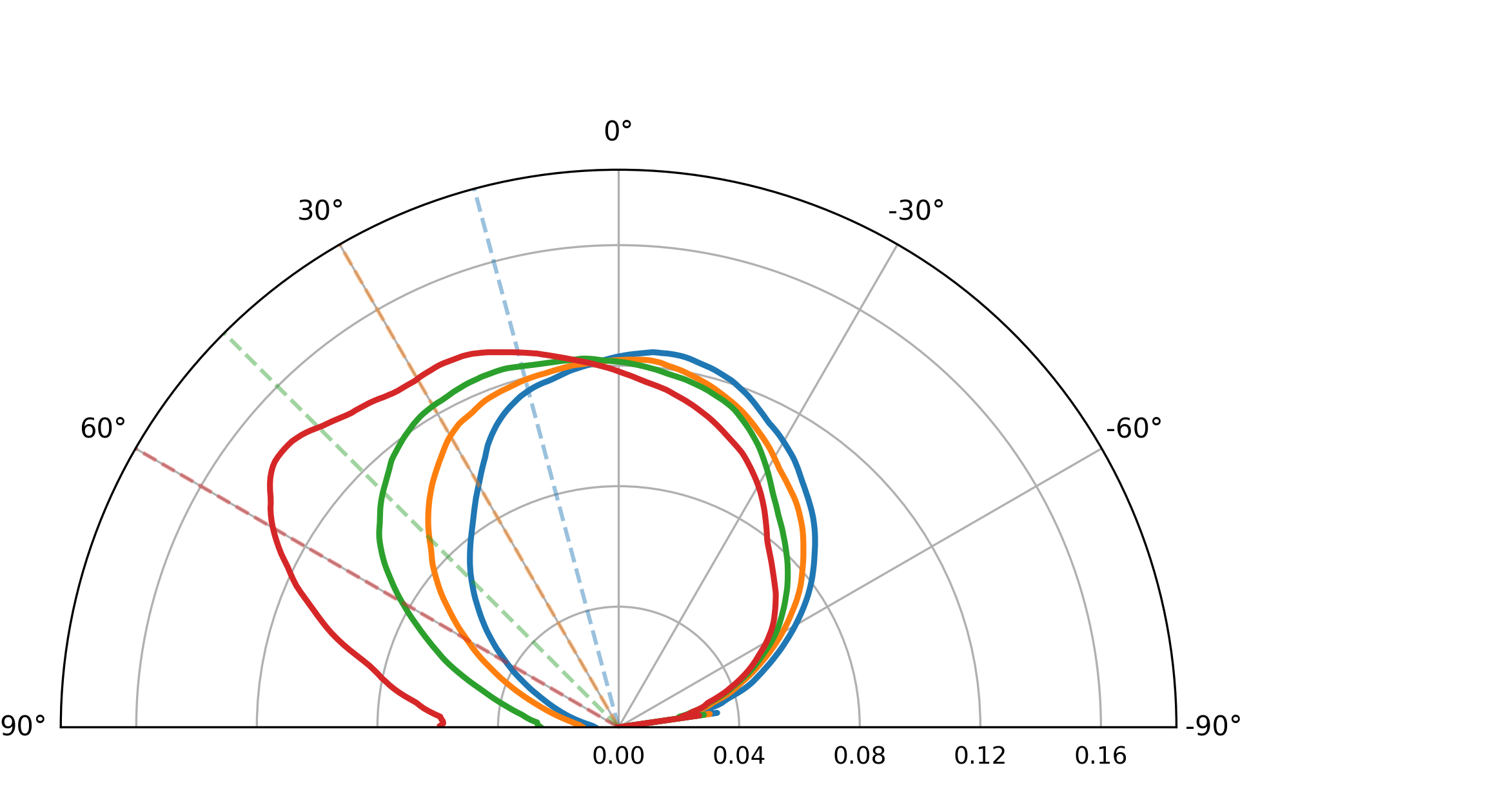} &
  \includegraphics[width=0.235\linewidth]{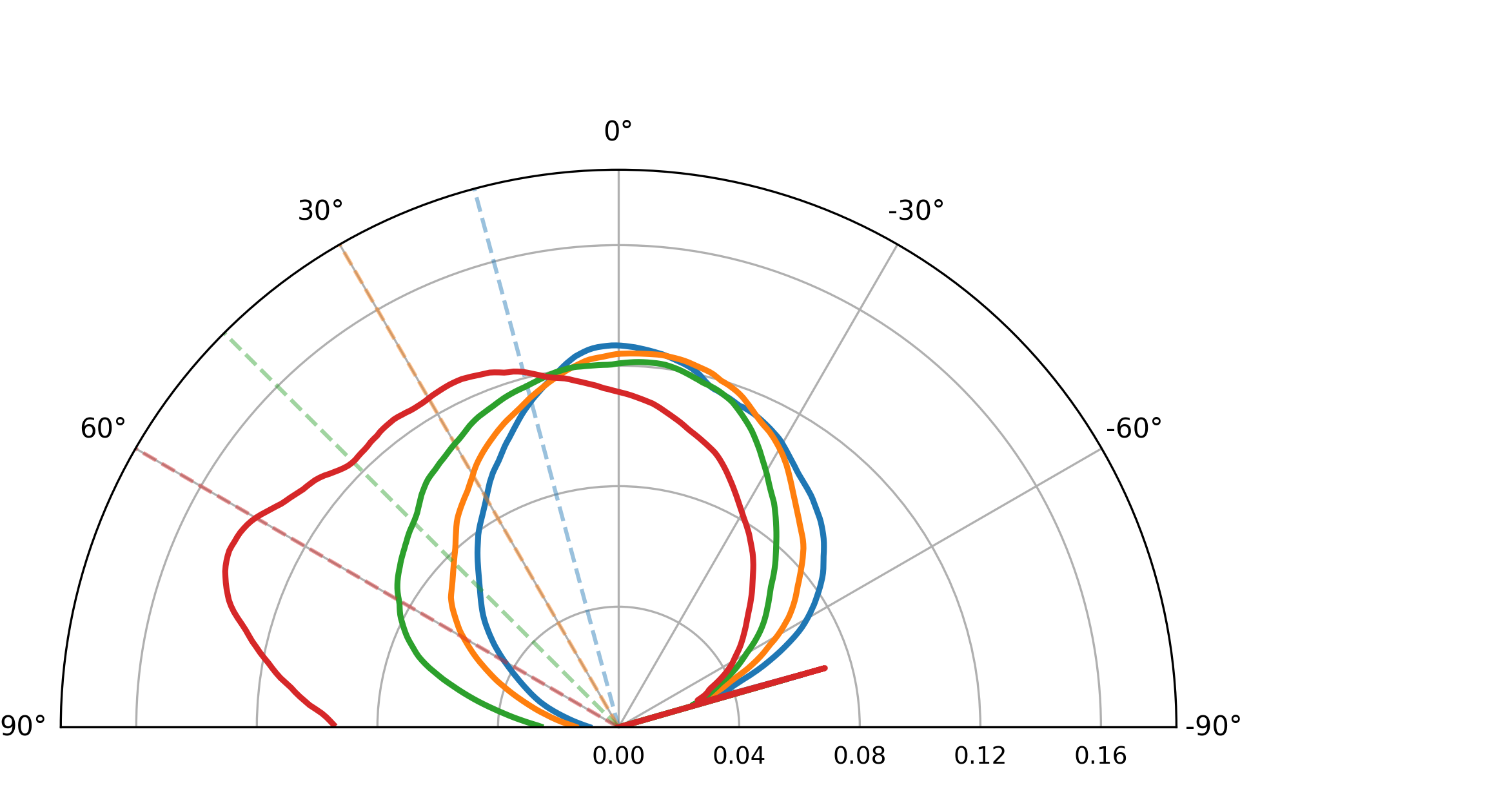} &
  \includegraphics[width=0.235\linewidth]{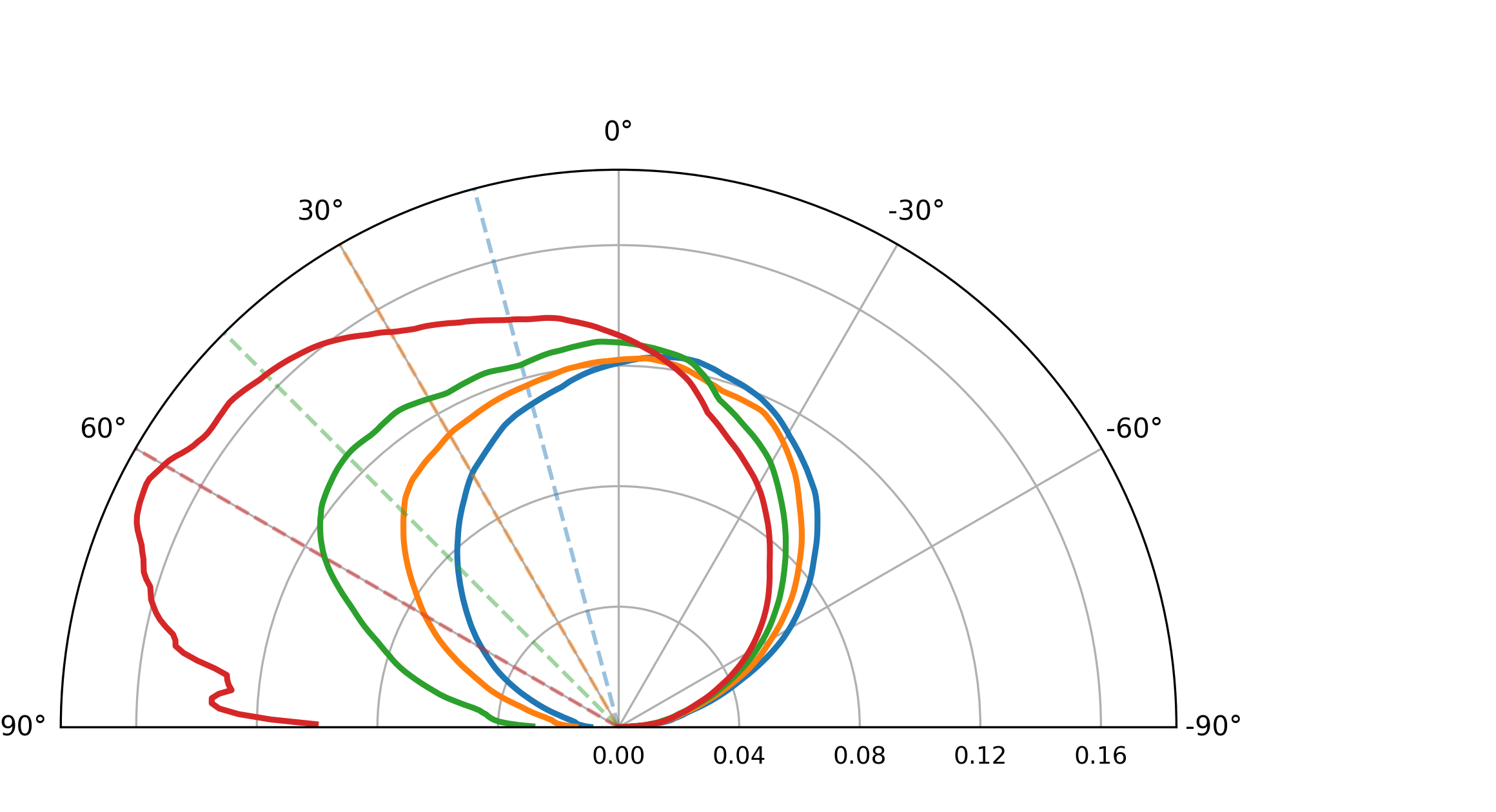} \\
  \includegraphics[width=0.235\linewidth]{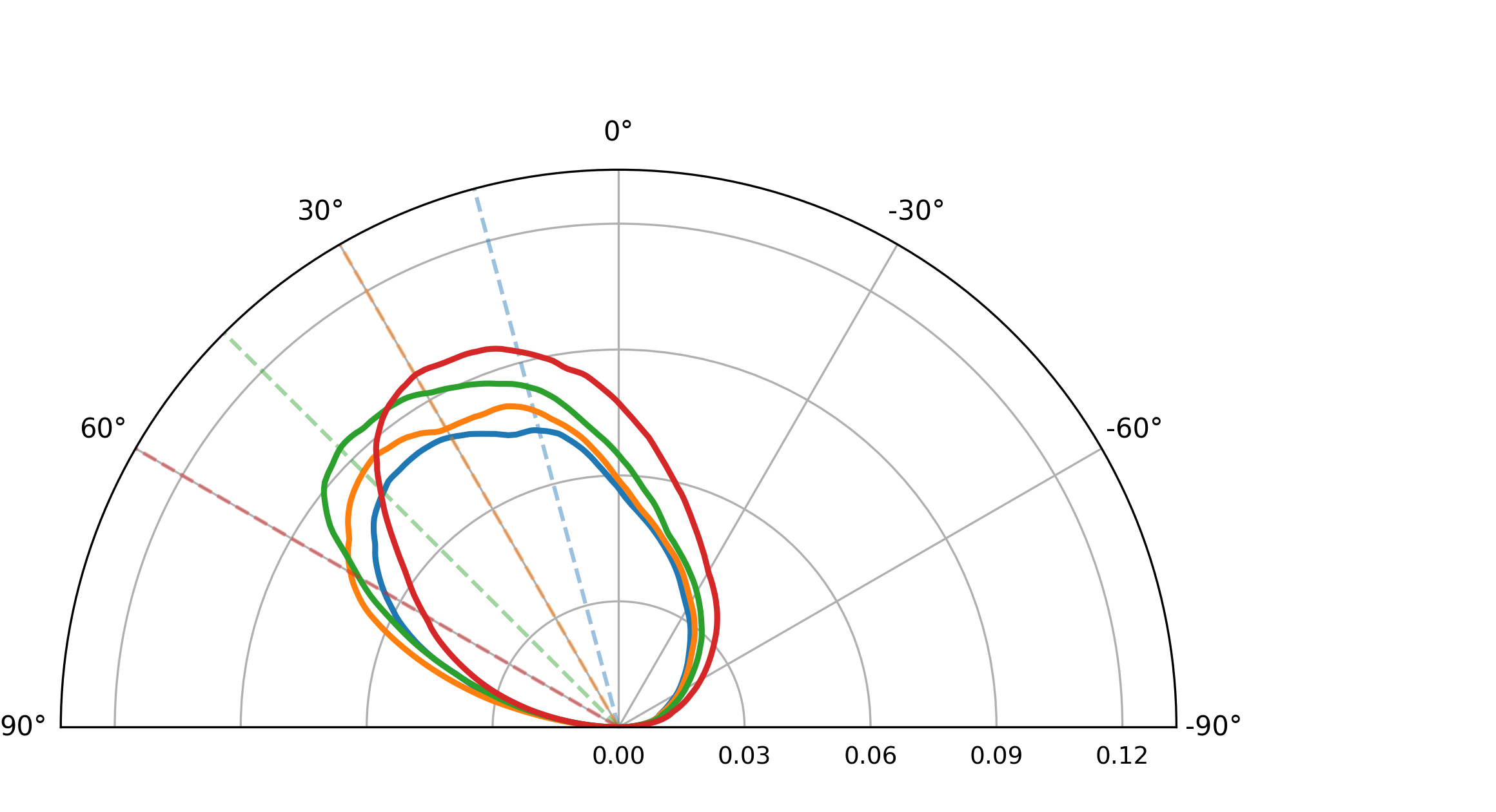} &
  \includegraphics[width=0.235\linewidth]{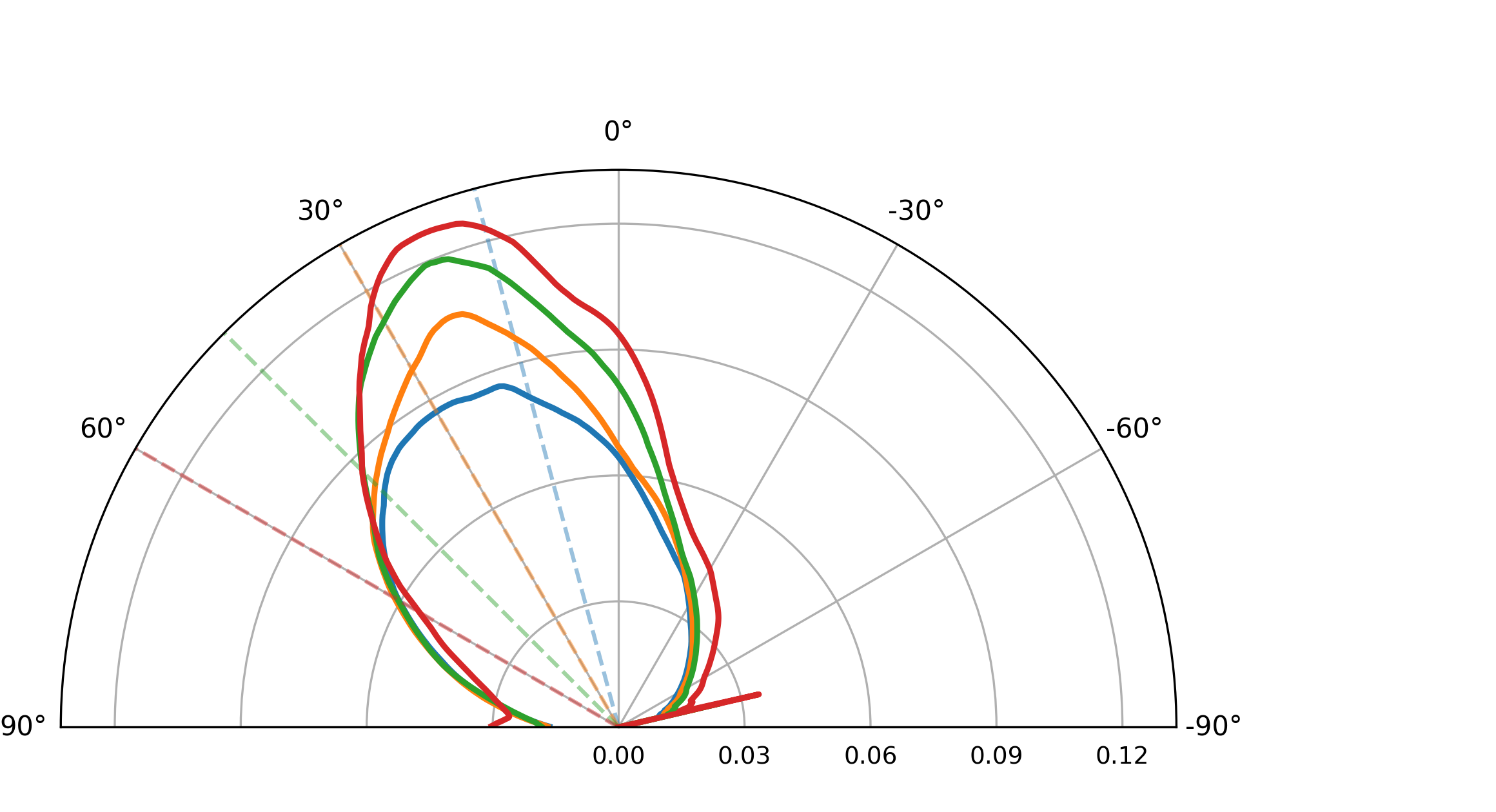} &
  \includegraphics[width=0.235\linewidth]{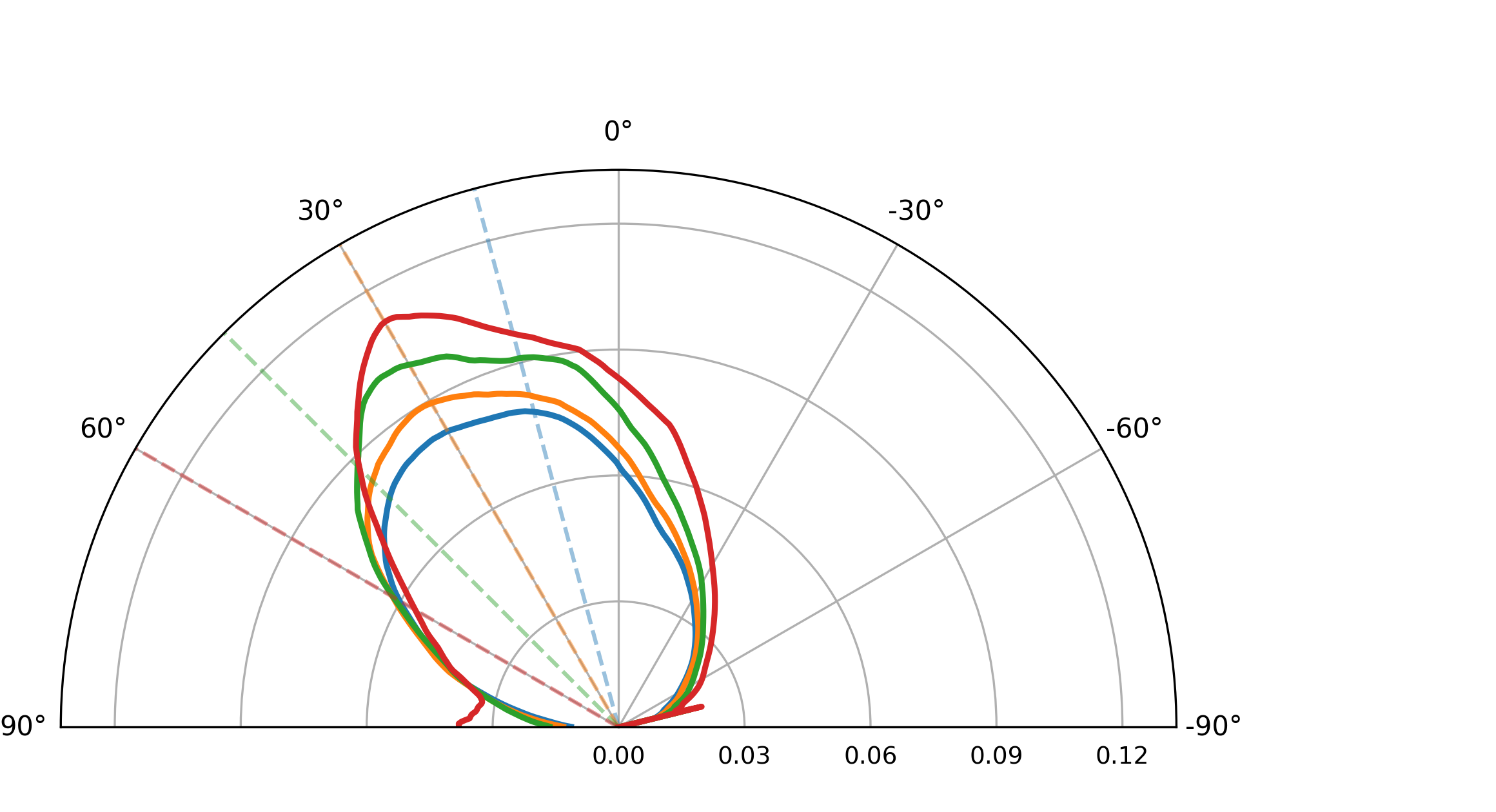} &
  \includegraphics[width=0.235\linewidth]{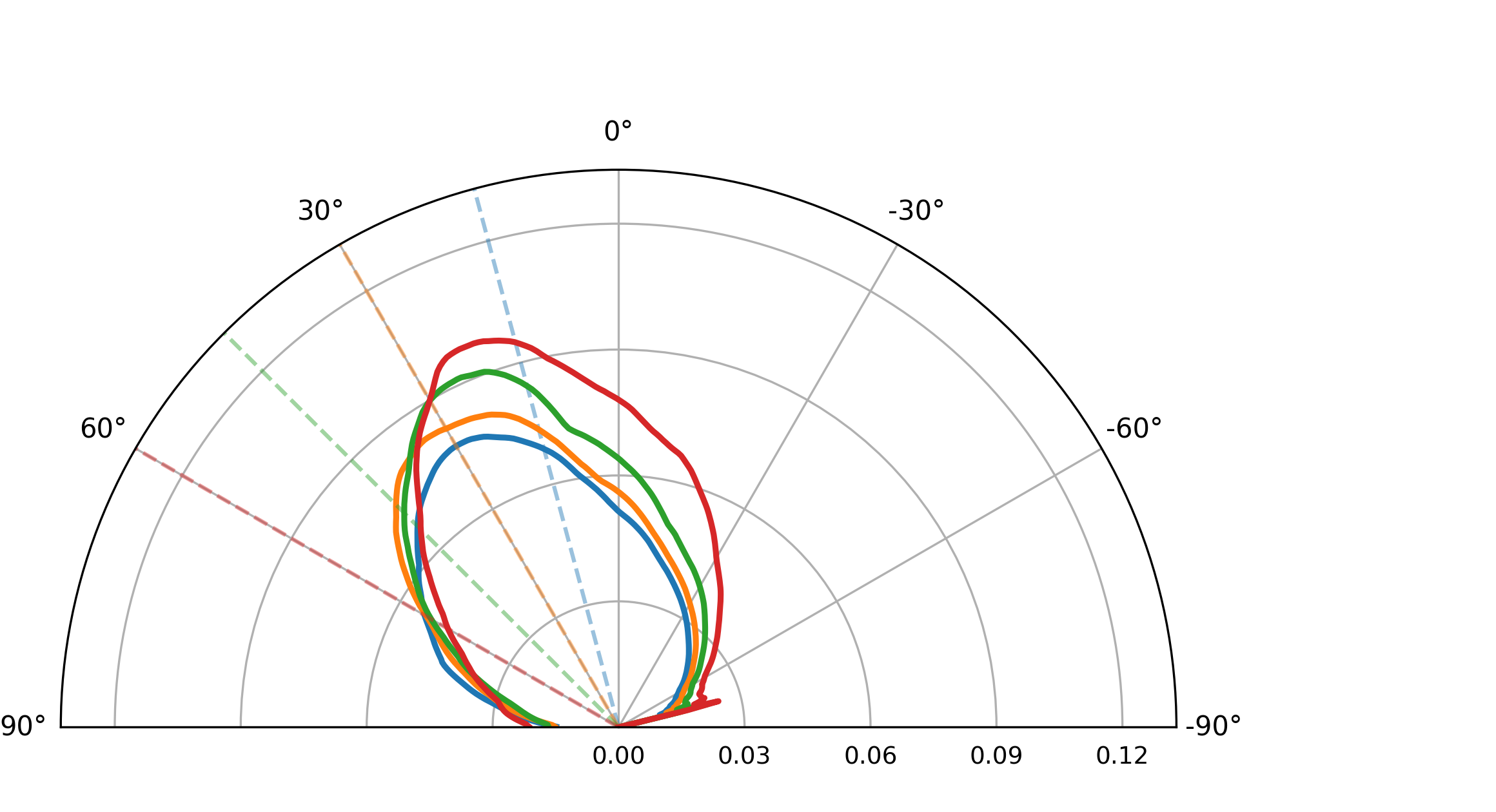} \\
  \includegraphics[width=0.235\linewidth]{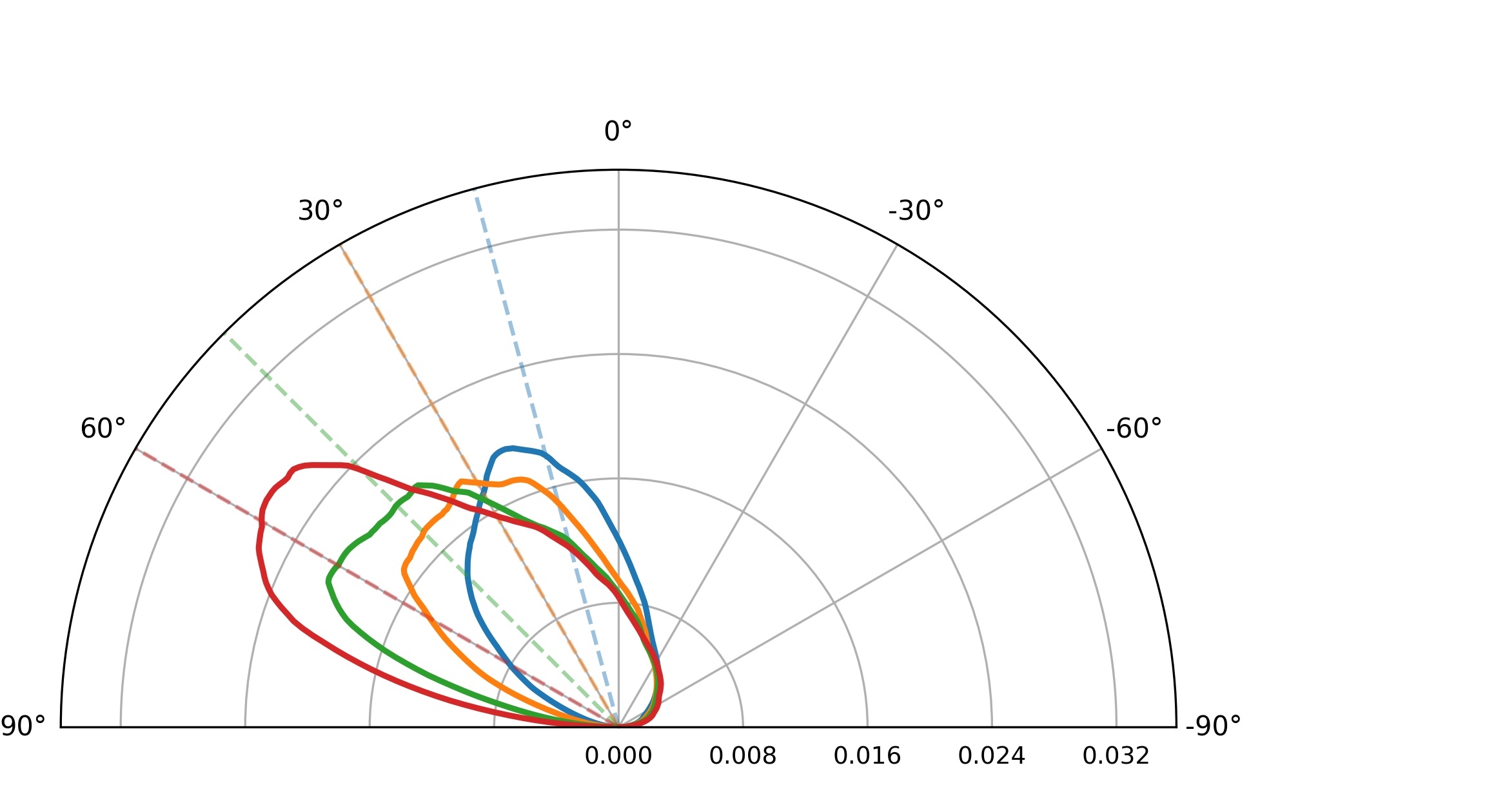} &
  \includegraphics[width=0.235\linewidth]{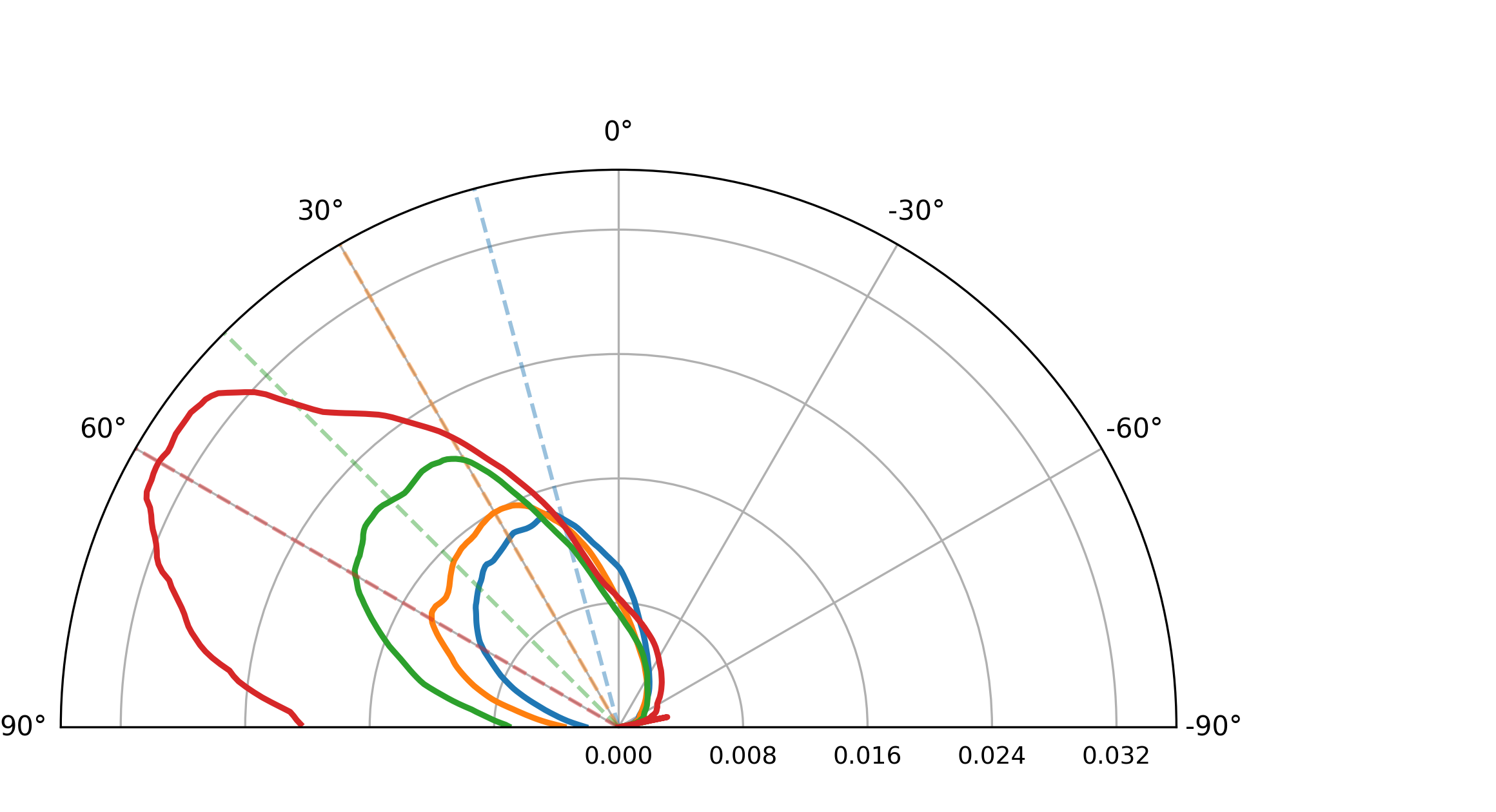} &
  \includegraphics[width=0.235\linewidth]{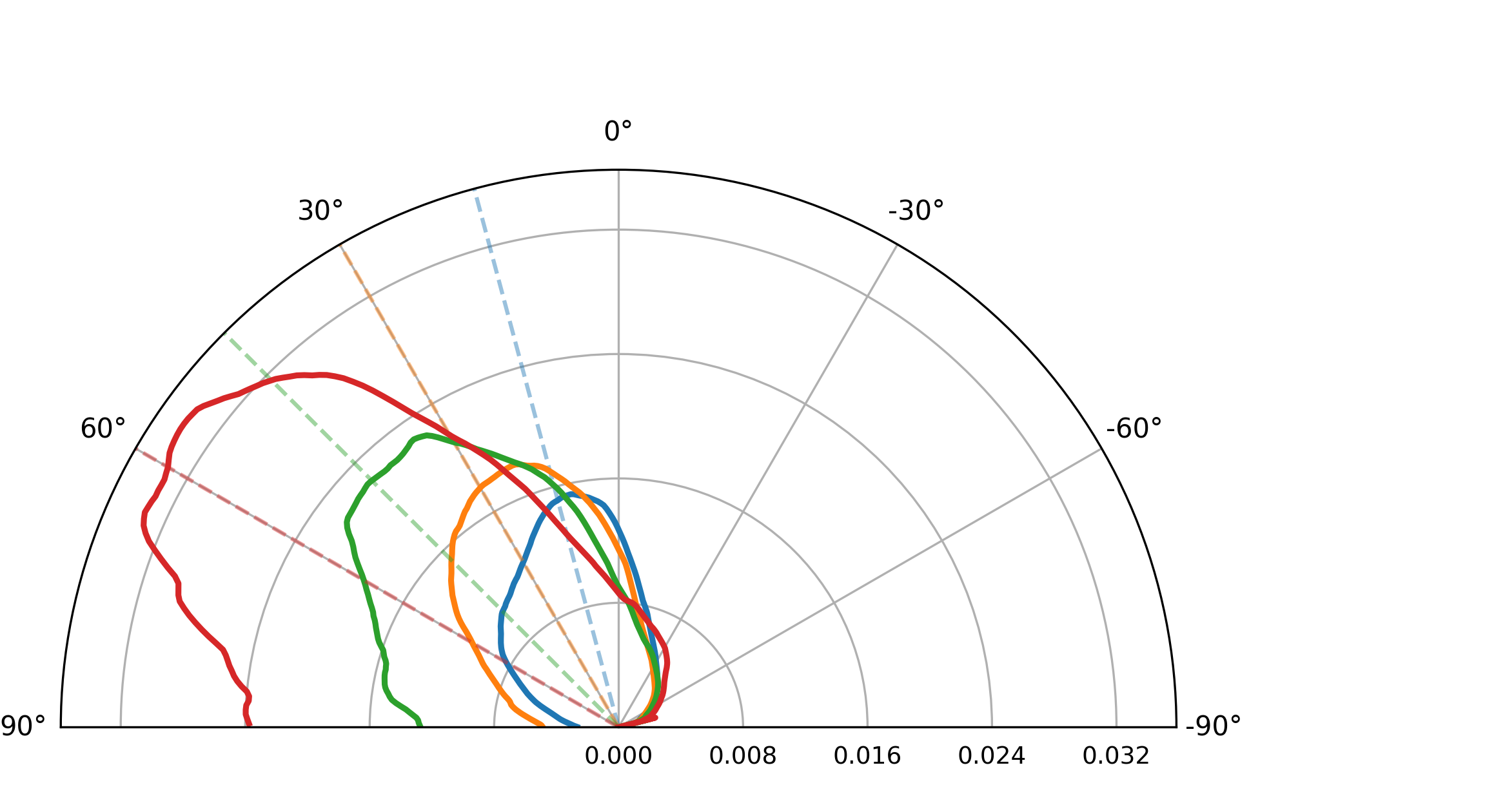} &
  \includegraphics[width=0.235\linewidth]{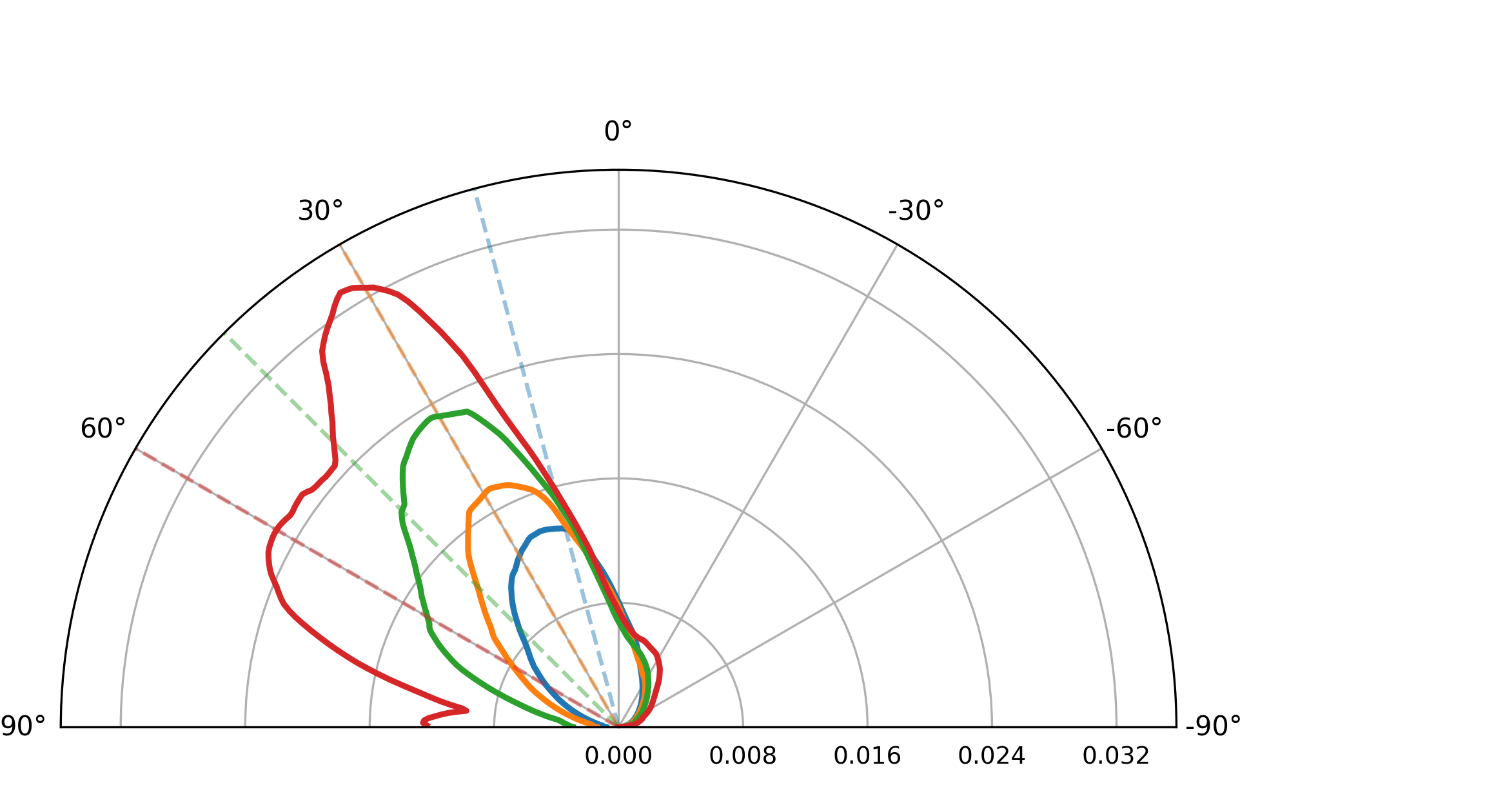} \\
  \end{tabular}
  \caption{\textbf{Grazing-angle supervision.} Polar BRDF-lobe plots of the learned contribution $f_r\cos\theta_i$ vs.\ incident elevation $\theta_i$ ($0^\circ$=normal at top, $\pm90^\circ$=grazing at the sides); each colored curve is a fixed outgoing direction (blue~$\theta_o\!=\!15^\circ$, orange~$30^\circ$, green~$45^\circ$, red~$60^\circ$), and the dashed line marks the specular direction. Each row is a different surface point; columns are the four grazing-supervision variants. Only our loss (first column, Eq.~\ref{eq:decay}) makes $f_r\cos\theta_i$ fall off smoothly to zero near grazing; ZeroExact, NearZeroBRDF and the no-grazing baseline all leave the near-grazing region uncontrolled, inflating the lobes toward $\pm90^\circ$.}
  \label{fig:ablation_grazing}
\end{figure*}

To visualize the learned reflectance we use the Stage-1 BRDF-lobe plot: we
select one learned point latent, freeze the decoder,
fix the outgoing direction $\boldsymbol{\omega}_o$ (one colored curve per
$\theta_o\!\in\!\{15,30,45,60\}^\circ$), and sweep the incident elevation
$\theta_i$ from $0^\circ$ (surface normal) to $\pm90^\circ$ (grazing) at a
fixed azimuth; the polar radius is the contribution $f_r\cos\theta_i$ and the
dashed line marks the specular direction. \Cref{fig:ablation_grazing}
shows four representative surface points.

Only our loss (\cref{eq:decay}) yields lobes that taper smoothly to zero
at grazing. \textbf{ZeroExact} fixes only the single, measure-zero point
$\theta_i\!=\!90^\circ$ and leaves its entire neighborhood free, so the
contribution balloons outward toward grazing with no smooth transition;
forcing $f_r$ itself to zero is also unphysical, since the raw BRDF is
generally non-zero near grazing --- it is the \emph{product} $f_r\cos\theta_i$
that must vanish. \textbf{NearZeroBRDF} pins $f_r$ to zero across a thin
near-grazing band but cannot shape the transition between the last supervised
observation and the grazing region, leaving non-physical lobes and spikes
near $\pm90^\circ$. The no-grazing baseline is unconstrained and shows the
same inflation.

\section{Ablation: Dataset Size}
\label{supp:datasize}

We ablate the size of the Stage-1 training corpus to test whether our
dataset provides a sufficiently diverse distribution that the learned
prior keeps improving as more materials are added. We draw random
subsets of $100$, $300$, and $500$ cloth materials and apply the same
$0.8/0.2$ split to each, yielding $80$, $240$, and $400$ training
materials respectively (denoted RoboCloth-100/300/500). Every subset trains a
decoder for the \emph{same} number of Stage-1 iterations
($4.8\times10^5$ steps) under the identical architecture and loss, and
is then evaluated with the Stage-2 UBO2014 transfer protocol of
\Cref{sec:results}: we freeze the decoder and optimize only the
per-point latent and the per-channel scale $\boldsymbol{\beta}$ on the
same $12$ UBO2014 materials for $60$ epochs.

\Cref{fig:datasize} and \Cref{tab:datasize} report the mean
UBO2014 PSNR over all $12$ transfer materials.
Transfer PSNR increases monotonically with the training-set size
($36.28\rightarrow36.57\rightarrow36.76$~dB), and the per-material
numbers improve almost everywhere: the decoder consistently benefits
from more of our data, indicating that the corpus carries genuine
appearance diversity rather than redundant samples. The gain also
flattens toward the full set ($+0.29$~dB from $80$ to $240$ materials
versus $+0.19$~dB from $240$ to $400$), suggesting that our full corpus
is approaching saturation for this transfer task.

\input{figures/datasize}

\begin{table}[t]
\centering\small
\setlength{\tabcolsep}{6pt}
\begin{tabular}{l ccc}
\toprule
Material & RoboCloth-100 & RoboCloth-300 & RoboCloth-500 \\
         & {\scriptsize(80)} & {\scriptsize(240)} & {\scriptsize(400)} \\
\midrule
fabric02 & 34.38 & \textbf{34.60} & \underline{34.57} \\
fabric04 & 33.33 & \underline{33.72} & \textbf{34.00} \\
fabric09 & 38.93 & \underline{39.40} & \textbf{39.50} \\
fabric11 & 33.76 & \underline{34.04} & \textbf{34.33} \\
felt01   & 38.38 & \underline{38.70} & \textbf{38.76} \\
felt03   & 39.75 & \underline{40.29} & \textbf{40.30} \\
felt05   & 34.82 & \underline{35.05} & \textbf{35.39} \\
felt10   & 43.62 & \underline{43.91} & \textbf{44.36} \\
carpet02 & 38.22 & \underline{38.58} & \textbf{38.87} \\
carpet07 & 34.22 & \underline{34.37} & \textbf{34.61} \\
carpet09 & 32.97 & \underline{33.05} & \textbf{33.17} \\
carpet12 & 33.01 & \underline{33.11} & \textbf{33.25} \\
\midrule
\textbf{Average} & 36.28 & \underline{36.57} & \textbf{36.76} \\
\bottomrule
\end{tabular}
\caption{\textbf{Dataset-size ablation (UBO2014 transfer).}
Per-material PSNR (dB,~$\uparrow$) for decoders trained on the
$100/300/500$-material subsets ($80/240/400$ training materials after the
$0.8/0.2$ split). PSNR rises consistently with more training data.}
\label{tab:datasize}
\end{table}

\section{Dataset Construction: Full Details}
\label{supp:dataset}

\subsection{Capture Trajectory}
\label{supp:trajectory}

The camera viewpoints are drawn from a hemisphere above the sample,
with polar angle $\phi \in [0, 2\pi/5]$ and two azimuthal lobes at
$\theta = 0$ and $\theta = \pi$. The hemisphere is partitioned into
8 sectors visited in a space-filling order to minimize arm travel.
Light positions are sampled by ray-bounding-box intersection on a
wireframe cage above the sample: $N_\text{base} = 50$ base positions
cover the upper hemisphere uniformly, and $N_\text{tail} = 8$
additional positions provide finer angular resolution near grazing.
Between consecutive frames the turntable advances by $29 \pm 19$
degrees, with a uniform $[0, 360)$ initial rotation drawn once per
material. Reproducibility is ensured by a \texttt{NumPy SeedSequence}
keyed on the cloth identifier: light positions are shared across all
materials in a 10-batch, while turntable jitter and per-step camera
$\phi$ jitter are unique per material.

\subsection{Hand-Eye and Turntable-Axis Calibration}
\label{supp:calibration}

\paragraph{Hand-eye calibration.}
The hand-eye transform ${}^{g}T_c$ is recovered from a
sequence of ChArUco-board~\cite{garrido2014automatic,romero2018speeded}
observations using Tsai's
formulation~\cite{tsai1989new}, as implemented in
\texttt{cv2.calibrateHandEye}. For every frame the board pose
${}^{c}T_{t,i}$ is estimated with \texttt{solvePnP} from the detected
board corners; pairing these target-to-camera poses with the robot
gripper-to-base poses $\{{}^{b}T_{g,i}\}$ gives the $AX = XB$ system
that we solve for ${}^{g}T_c$. The calibration is
performed once per camera mounting and stored as a fixed transform for
all subsequent captures.

\paragraph{Turntable-axis calibration.}
The turntable rotation axis $(\hat{\mathbf{n}}, \mathbf{c})$ is
calibrated from a capture scan itself, by aligning the COLMAP camera
centers $\{\mathbf{p}_i\}$ with the robot-derived camera centers
$\{\mathbf{q}_i\}$ obtained from the logged gripper poses and the
hand-eye transform. Writing $R_i = R(\hat{\mathbf{n}}, \theta_i)$ for
the rotation by the logged turntable angle $\theta_i$, we alternate
three steps to convergence: (i) a Umeyama similarity
fit~\cite{umeyama1991least} of $\{\mathbf{q}_i\}$ onto the centers
predicted by the current $(\hat{\mathbf{n}}, \mathbf{c})$, giving
$(s, R, \mathbf{t})$; (ii) a linear least-squares update of the axis
point from the stacked system
$(I - R_i)\,\mathbf{c} = sR\,\mathbf{q}_i + \mathbf{t} - R_i\,\mathbf{p}_i$;
and (iii) a Levenberg--Marquardt step on $\hat{\mathbf{n}}$,
parameterized in its tangent space so that it stays a unit vector. A
final polish freezes the similarity and refines
$(\hat{\mathbf{n}}, \mathbf{c})$ alone under a Huber loss. Turntable
angles are counter-clockwise about $\hat{\mathbf{n}}$ by the
right-hand rule, which fixes the
$\hat{\mathbf{n}} \equiv -\hat{\mathbf{n}}$ sign convention.

\paragraph{Derotation derivation.}
For completeness we derive the derotation operator used in
\Cref{sec:geometry}. Let $R(\hat{\mathbf{n}}, \alpha)$ denote
the Rodrigues rotation by angle $\alpha$ about $\hat{\mathbf{n}}$, and
define
\[
\mathcal{T}(R, \mathbf{c})
=
\begin{bmatrix} R & (I - R)\mathbf{c} \\ \mathbf{0} & 1 \end{bmatrix}
\in \mathrm{SE}(3),
\]
which rotates by $R$ about the point $\mathbf{c}$. The derotation
transform that undoes turntable angle $\theta_i$ is
\[
{}^{w_0}T_{b,i}
= \mathcal{T}\!\bigl(R(\hat{\mathbf{n}}, -\theta_i),\, \mathbf{c}\bigr),
\]
and the camera pose in the common $0^\circ$ world frame is
\[
{}^{w_0}T_{c,i}
= {}^{w_0}T_{b,i} \, {}^{b}T_{g,i} \, {}^{g}T_c.
\]

\subsection{Similarity Alignment: Umeyama Derivation}
\label{supp:umeyama}

Let $\{\mathbf{p}_i\}$ be the COLMAP camera centers and
$\{\mathbf{q}_i\}$ the corresponding robot-derived centers in the
$0^\circ$ frame. We solve for $(s, R, \mathbf{t}) \in \mathrm{Sim}(3)$
such that $\mathbf{q}_i \approx sR\mathbf{p}_i + \mathbf{t}$. Define
\[
\bar{\mathbf{p}} = \tfrac{1}{N}\textstyle\sum_i \mathbf{p}_i,
\quad
\bar{\mathbf{q}} = \tfrac{1}{N}\textstyle\sum_i \mathbf{q}_i,
\quad
\Sigma = \tfrac{1}{N}\textstyle\sum_i
(\mathbf{q}_i - \bar{\mathbf{q}})(\mathbf{p}_i - \bar{\mathbf{p}})^\top.
\]
The SVD $\Sigma = UDV^\top$ yields
\[
\begin{gathered}
R = USV^\top,
\quad
S = \mathrm{diag}(1, 1, \det(UV^\top)), \\
s = \frac{\mathrm{tr}(DS)}{\frac{1}{N}\sum_i \|\mathbf{p}_i -
\bar{\mathbf{p}}\|^2},
\quad
\mathbf{t} = \bar{\mathbf{q}} - sR\bar{\mathbf{p}}.
\end{gathered}
\]
The correction matrix $S$ handles the reflection ambiguity in the SVD.
The mean alignment residual
$\frac{1}{N}\sum_i \|sR\mathbf{p}_i + \mathbf{t} - \mathbf{q}_i\|$
is logged per material as a reconstruction quality diagnostic; frames
with per-image residual exceeding $\tau = 16\,\mathrm{mm}$ are
discarded before the observation tensor is assembled.

\subsection{Radiometric Calibration Details}
\label{supp:radiometric}

\paragraph{Demosaicing and quantum-efficiency correction.}
The Spinnaker sensor outputs a 14-bit Bayer mosaic in RGGB pattern,
stored left-aligned in a 16-bit PNG. We demosaic using the
Menon~2007 algorithm~\cite{menon2007demosaicing} and apply a
channel-wise quantum-efficiency correction
$(g_R, g_G, g_B) = (1.5806, 1.0, 1.0659)$ before clipping to the
16-bit range. Both the raw mosaic and the debayered HDR PNG are
retained on disk.

\paragraph{Camera factor calibration.}
The camera factor $\kappa$ converts pixel values to physical radiance
under the linear sensor model. It is measured once per exposure cohort
by imaging a near-Lambertian white reference card while sweeping the
LED through a fixed polar grid (\texttt{linear\_hemi} routine). The
predicted irradiance at each step is known analytically from the robot
pose and the emitter model $\Lambda(\theta)$; fitting the slope of
measured intensity against predicted radiance yields $\kappa$. For our
\texttt{Dataset\_Nov11} captures, three exposure cohorts
$\{20000, 20000, 8000\}\,\mu\mathrm{s}$ give factors
$(\kappa_1, \kappa_2, \kappa_3)$ satisfying
$\kappa_3 = \kappa_2 \cdot (8000/20000)$ to within the fit residual.

\paragraph{Emitter directional distribution.}
For every observation of the white reference card we compute the ratio
$r = I_\text{pred}^\text{iso} / I_\text{meas}$, bin by the
LED-emission angle $\theta$ in $1^\circ$ bins over
$[0^\circ, 90^\circ]$, average within each bin, fill empty bins by
nearest-neighbor interpolation, and normalize by the maximum. The
result is a single 90-entry table $\Lambda(\theta) \in [0,1]$, shared
across the three colour channels.

\paragraph{Color correction matrix.}
A $3\times3$ CCM is estimated from a 24-patch ColorChecker.
Reference values are converted from manufacturer Lab$(D_{50})$ to
linear sRGB$(D_{65})$, where $D_{50}$ and $D_{65}$ are the two spaces' reference white illuminants, via
\[
\begin{gathered}
\mathrm{Lab}(D_{50})
\rightarrow \mathrm{XYZ}(D_{50})
\xrightarrow{\text{Bradford CAT}} \\
\mathrm{XYZ}(D_{65})
\rightarrow \text{linear sRGB}(D_{65}),
\end{gathered}
\]
and the matrix $M = \arg\min_M \|CM - R\|_F^2$ is solved in closed
form, where $C \in \mathbb{R}^{24\times3}$ holds captured patch colors
and $R \in \mathbb{R}^{24\times3}$ the references. Color accuracy is
assessed with the CIEDE2000 color difference $\Delta E_{00}$~\cite{luo2001development} on the same 24 patches. The CCM is
shipped alongside the data but not applied to the released images;
see \Cref{sec:radiometric} for the rationale.

\section{Pipeline Module Details}
\label{supp:modules}

\begin{figure}[t]
  \centering
  \includegraphics[width=\linewidth]{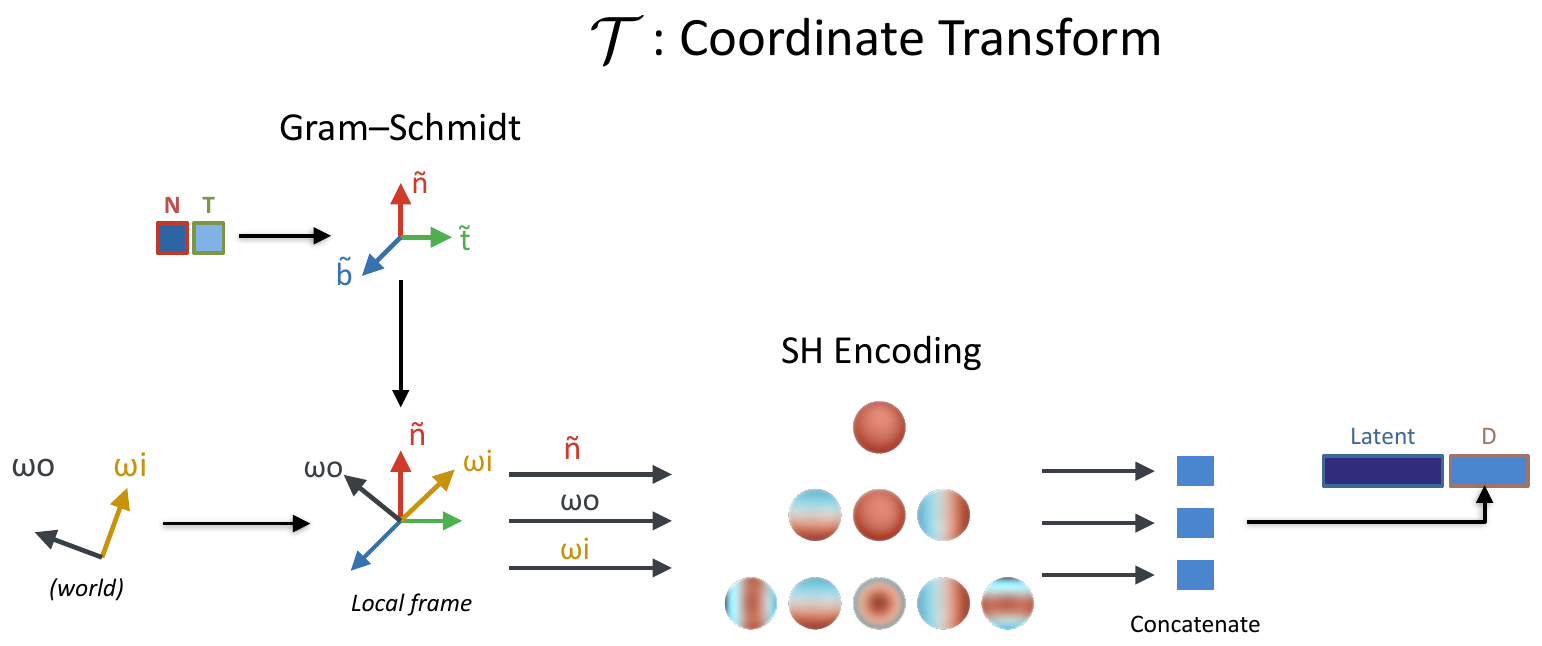}
  \caption{\textbf{The coordinate-transform operator $\mathcal{T}$.}
    A per-point (Stage~1) or per-texel (Stage~2) normal $\mathbf{n}$ and
    tangent $\mathbf{t}$ are orthogonalized by Gram--Schmidt into a TBN
    frame $(\tilde{\mathbf{n}},\tilde{\mathbf{t}},\tilde{\mathbf{b}})$.
    The world-space incident and outgoing directions
    $\boldsymbol{\omega}_i,\boldsymbol{\omega}_o$ are rotated into this
    local frame; the local directions together with $\tilde{\mathbf{n}}$
    are then encoded with degree-$3$ real spherical harmonics and
    concatenated with the latent code before being passed to the BRDF
    decoder.}
  \label{fig:module_transform}
\end{figure}

\begin{figure}[t]
  \centering
  \includegraphics[width=\linewidth]{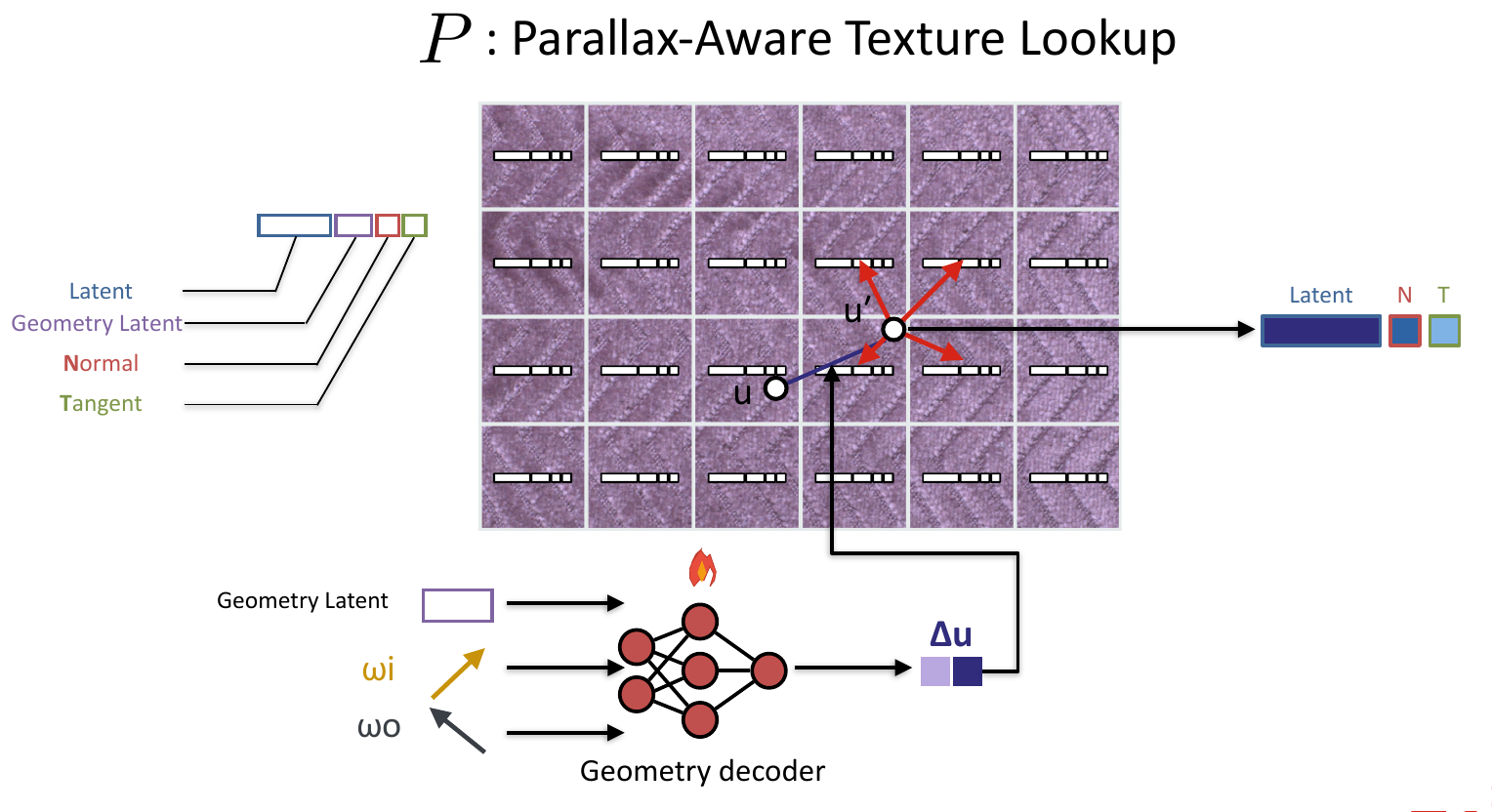}
  \caption{\textbf{Parallax-aware texture lookup $P$ (Stage~2).}
    An initial sample at $\mathbf{u}$ provides the geometry latent
    $\mathbf{g}(\mathbf{u})$. Together with the incident and outgoing
    directions $\boldsymbol{\omega}_i,\boldsymbol{\omega}_o$, a small
    geometry decoder maps it to a texture-space displacement
    $\Delta\mathbf{u}$. The final point attributes are bilinearly sampled
    from the latent-geometry grid $\mathbf{Z}$ at
    $\mathbf{u}'=\mathbf{u}+\Delta\mathbf{u}$, reproducing yarn-level
    parallax without an explicit height field.}
  \label{fig:module_lookup}
\end{figure}

The two operators that appear as labeled boxes in the main-paper
pipeline (Figs.~\ref{fig:stage1} and~\ref{fig:stage2}) are expanded here: the
coordinate transform $\mathcal{T}$ (\Cref{fig:module_transform}),
shared by both stages, and the parallax-aware texture lookup $P$
(\Cref{fig:module_lookup}), used in Stage~2.

\noindent\textbf{Construction of $\mathcal{T}$.}
Given a point's normal $\mathbf{n}$ and tangent $\mathbf{t}$, $\mathcal{T}$ orthonormalizes a TBN frame by Gram--Schmidt,
\[
\tilde{\mathbf{n}} = \hat{\mathbf{n}},\quad
\tilde{\mathbf{t}} = \widehat{\mathbf{t} - (\mathbf{t}\cdot\tilde{\mathbf{n}})\,\tilde{\mathbf{n}}},\quad
\tilde{\mathbf{b}} = \tilde{\mathbf{n}}\times\tilde{\mathbf{t}},
\]
rotates the incident and outgoing directions into this frame with $R=[\,\tilde{\mathbf{t}}\;\tilde{\mathbf{b}}\;\tilde{\mathbf{n}}\,]$, and encodes the rotated directions together with $\tilde{\mathbf{n}}$ by degree-$3$ real spherical harmonics ($48$ dims), concatenating the latent code $\mathbf{z}$ to form the decoder input. The orthogonalization is differentiable, so gradients flow back into the unconstrained $(\mathbf{n},\mathbf{t})$ tables without a projected-gradient step.

\noindent\textbf{Construction of $P$.}
In Stage~2, let $[\cdot]_{\mathrm{geom}}$ select the geometry channels of
the latent-geometry grid. The initial lookup and learned displacement are
\[
\begin{aligned}
\mathbf{g}(\mathbf{u})
  &= \bigl[\mathrm{BilinearSample}(\mathbf{Z},\mathbf{u})\bigr]_{\mathrm{geom}},\\
\Delta\mathbf{u}
  &= g_\phi\!\left(\mathbf{g}(\mathbf{u}),\boldsymbol{\omega}_i,
     \boldsymbol{\omega}_o\right)\in\mathbb{R}^2.
\end{aligned}
\]
The parallax-aware texture lookup then returns the complete point attributes
from the displaced coordinate,
\[
P(\mathbf{Z},\mathbf{u},\boldsymbol{\omega}_i,\boldsymbol{\omega}_o)
= \mathrm{BilinearSample}\!\bigl(\mathbf{Z},\ \mathbf{u}+\Delta\mathbf{u}\bigr).
\]
The displacement shifts the sampled texel with view and light so the
per-texel attributes follow the visible yarn surface; setting
$\Delta\mathbf{u}=\mathbf{0}$ recovers a standard bilinear lookup.

\subsection{Stage-1 Training Sources}
\label{supp:training_sources}

\begin{table}[t]
\small
\centering
\begin{tabular}{l l c}
\toprule
Source & Data type & \# Materials \\
\midrule
RoboCloth  & Real robotic captures (HDR)  & $\sim$400 \\
Bonn       & Multi-LED and linear-light-source & $311$ \\
MERL       & Measured isotropic BRDFs     & $100$ \\
\bottomrule
\end{tabular}
\caption{Training sources for the neural decoder (Stage~1). Each
RoboCloth material contributes ${\sim}580$ HDR frames. Every Bonn
material is measured in three modalities: PAN (panchromatic, $24$
unfiltered LEDs, $388$ images), POLY (polychromatic, RGB-filtered
LEDs, $100$ images), and LLS (linear light source, $14$ angles, $280$
images). MERL supplies tabulated BRDFs rather than images and is
sampled directly.}
\label{tab:training_sources}
\end{table}

\textbf{RoboCloth} is loaded from the structured $(K, V, 3)$ \texttt{uint16}
tensors described in \Cref{sec:obs_tensor}. All retained
materials are held in CPU memory and sampled uniformly across (material,
point, view) triples.
\textbf{Bonn}~\cite{merzbach2019learned} is loaded from the released AxF
files. Here the ``Bonn'' source is UBOFAB19~\cite{merzbach2019learned},
the fabric release (v1) of the Bonn SVBRDF dataset; we use ``Bonn'' as
shorthand for it throughout.
PAN and POLY observations are fed to the renderer with a single
emitter sample ($S\!=\!1$) using the published per-LED point-light
positions. LLS observations are integrated with $S\!=\!4$ Monte Carlo
samples drawn from the four corners of each linear light source.
\textbf{MERL}~\cite{matusik2003data} is a real, measured source: we render
each of the $100$ isotropic BRDFs from the Rusinkiewicz-parameterized
table at random $(\theta_h, \theta_d, \phi_d)$ triples. The renderer
degenerates to a direct BRDF lookup — no emitter, no inverse-square
fall-off — and the loss reduces to a direct fit on tabulated values.

When these sparse sources (and the UBO2014 test set) are used as
Stage~2 reconstruction targets, they are fit with the same per-point
latent representation as Stage~1---one
$(\mathbf{z},\mathbf{n},\mathbf{t})$ triplet per observation point,
optimized with the frozen decoder and the per-channel scale
$\boldsymbol{\beta}$---rather than the dense latent-geometry grid and
parallax-aware texture lookup $P$, which require the continuously
parametrized surface available only for our own captures.

\section{Disney Principled PBR Baseline}
\label{supp:pbr}

The non-neural reference of \Cref{sec:datasets} fits, per material, a
set of Disney principled BRDF~\cite{burley2012physically} parameter maps with the
same renderer and optimizer as Stage~2. At each shading point the BRDF is
the sum of an anisotropic microfacet specular lobe, a Burley diffuse lobe,
and a sheen lobe,
\begin{equation}
f_r^{\text{PBR}}
= f_{\text{spec}} + f_{\text{diff}} + f_{\text{sheen}},
\end{equation}
evaluated in the local TBN frame with half-vector
$\mathbf{h}=\widehat{\boldsymbol{\omega}_i+\boldsymbol{\omega}_o}$. We write
$\operatorname{lerp}(a,b,t)=(1-t)\,a+t\,b$.

\paragraph{Specular.}
The specular lobe is the standard anisotropic microfacet term
\begin{equation}
f_{\text{spec}}
=
\frac{D(\mathbf{h})\,G(\boldsymbol{\omega}_i,\boldsymbol{\omega}_o)\,
      F(\boldsymbol{\omega}_o\!\cdot\!\mathbf{h})}
     {4\,(\mathbf{n}\!\cdot\!\boldsymbol{\omega}_i)\,
        (\mathbf{n}\!\cdot\!\boldsymbol{\omega}_o)},
\end{equation}
where $D$ and $G$ are the anisotropic GGX normal-distribution and Smith
masking-shadowing terms, evaluated with directional roughnesses
$\alpha_x=\rho\,(1+a)$ and $\alpha_y=\rho\,(1-\tfrac{1}{2}a)$ for roughness
$\rho$ and anisotropy strength $a$, with the tangent frame rotated in-plane
by a learned anisotropy angle. $F$ is the Schlick Fresnel term
$F=C_{\text{spec}}+(1-C_{\text{spec}})\,(1-\boldsymbol{\omega}_o\!\cdot\!\mathbf{h})^5$
with a tinted dielectric reflectance
\begin{equation}
C_{\text{spec}}
=
\operatorname{lerp}\!\Bigl(
  F_0^{\text{d}}\,
  \operatorname{lerp}(\mathbf{1}, C_{\text{tint}}, s_{\text{spec}}),\;
  \mathbf{c},\;
  \mu
\Bigr),
\end{equation}
where $\mathbf{c}$ is the base color, $\mu$ the metallic factor,
$s_{\text{spec}}$ the specular tint,
$F_0^{\text{d}}=\bigl(\tfrac{\eta-1}{\eta+1}\bigr)^2$ the dielectric
reflectance for IOR $\eta$, and
$C_{\text{tint}}=\mathbf{c}/(0.3\,c_R+0.6\,c_G+0.1\,c_B)$ the
luminance-normalized hue.

\paragraph{Diffuse.}
The diffuse lobe is Burley's roughness-dependent model,
\begin{equation}
f_{\text{diff}}
=
\frac{\mathbf{c}\,(1-\mu)\,\text{ao}}{\pi}\,
F_d(\boldsymbol{\omega}_i)\,F_d(\boldsymbol{\omega}_o),
\end{equation}
with $F_d(\boldsymbol{\omega})=1+(F_{90}-1)\,(1-\mathbf{n}\!\cdot\!\boldsymbol{\omega})^5$,
$F_{90}=0.5+2\rho\,(\boldsymbol{\omega}_i\!\cdot\!\mathbf{h})^2$, and ambient
occlusion $\text{ao}$.

\paragraph{Sheen.}
The sheen lobe adds the characteristic fabric edge glow,
\begin{equation}
f_{\text{sheen}}
=
(1-\boldsymbol{\omega}_i\!\cdot\!\mathbf{h})^5\,
w_{\text{sheen}}\,(1-\mu)\,
\operatorname{lerp}(\mathbf{1}, C_{\text{tint}}, s_{\text{sheen}}),
\end{equation}
with sheen strength $w_{\text{sheen}}$ and sheen tint $s_{\text{sheen}}$.
The optimized parameters are per-texel maps of base color $\mathbf{c}$,
ambient occlusion, roughness, metallic, IOR, anisotropy (strength and
rotation), specular tint, sheen, and sheen tint, together with a normal
map, all at $2048\!\times\!2048$ to match the neural latent grid.

\section{Implementation Details}
\label{supp:impl}

Full hyperparameters for Stage~1 and Stage~2 are reported in
\Cref{tab:implementation_details}.
This section gives the full implementation.

\paragraph{Architecture and optimization.}
The decoder is a four-layer MLP (width $256$, \texttt{LeakyReLU}(0.2),
skip connection at layer~2, softplus output); incident and outgoing
directions use real spherical harmonics of degree $L=3$ and the latent
dimension is $d=24$. Stage~1 jointly trains the shared decoder and the
per-point parameter tables for all $\approx 400$ materials with 8-bit Adam;
Stage~2 freezes the decoder and fits each held-out material with
full-precision Adam.
Owing to compute and memory limits, every material is trained and
fit on a fixed random $10\%$ subsample of its surface points, applied
identically to all sources so the per-material budget is equal.
In every Stage-2 fit we additionally optimize a per-channel scale
$\boldsymbol{\beta}$ that aligns the frozen decoder to the target
capture's color space and radiometric calibration.

\paragraph{Per-point optimizable parameters.}
In Stage~1, each sparse surface point owns three directly optimizable
quantities stored as \texttt{requires\_grad=True} tensors: the latent
code $\mathbf{z}_{m,j}$, the surface normal $\mathbf{n}_{m,j}$, and
the tangent $\mathbf{t}_{m,j}$. These are updated by the same Adam
step as $\theta$, with the learning rate for $(\mathbf{n}, \mathbf{t})$
matching that of the latent tables ($2\times10^{-3}$). The
Gram--Schmidt orthogonalization is applied in the forward pass and is
fully differentiable, so gradients flow back into the unconstrained
$(\mathbf{n}, \mathbf{t})$ parameters without any projected-gradient
step.

\paragraph{8-bit Adam in Stage~1.}
Storing the complete per-point latent tables for ${\sim}400$ materials with
$V_m \in [10^3, 5\times10^3]$ points each requires keeping
${\sim}400 \times 5000 \times 24 \times 4\,\mathrm{bytes} \approx
192\,\mathrm{GB}$ of optimizer state in full precision. We use
\texttt{bitsandbytes} 8-bit Adam to reduce this by $4\times$, fitting
all optimizer states on a single node without offloading.

\paragraph{Round-robin material sampling.}
At each Stage~1 iteration, the active material set is rotated by a
deterministic round-robin so that every material is touched within a
fixed window of iterations. Within the active set, (point, view) pairs
are sampled without replacement until the batch of $5\times10^5$ rays
is filled.

\paragraph{Validation and checkpointing.}
At validation time, emitter samples are accumulated in chunks of
$S_\text{chunk} = 2$ to bound peak memory. We checkpoint every
$10^4$ steps and select the checkpoint at which the Bonn validation
loss plateaus (${\approx}\,4.8\times10^5$ steps) as the Stage~2
initialization for all cross-dataset comparisons.

\begin{table*}[t]
\centering
\begin{minipage}{\textwidth}
\centering
\resizebox{\textwidth}{!}{%
\begin{tabular}{l l}
\toprule
Component & Setting \\
\midrule
Direction encoding      & Real spherical harmonics, $L=3$, $48$ dims total \\
Latent dimension        & $d = 24$ \\
Decoder MLP             & $[256,256,256,256]$, LeakyReLU(0.2), softplus output, skip @ layer~2 \\
Stage~1 optimizer       & 8-bit Adam, $\eta_\text{lat}{=}2\times10^{-3}$,
                          $\eta_\text{dec}{=}2\times10^{-4}$ \\
Stage~1 batch           & $5\times10^5$ rays, round-robin material sampling \\
Stage~1 iterations      & ${\approx}\,4.8\times10^5$ steps \\
Stage~1 loss            & Log-relative; $\rho_\text{ref}$ set to each source's
                          median BRDF magnitude: $0.02$ (Bonn),
                          $0.057$ (MERL), $1.5\times10^4$ (RoboCloth) \\
Stage~1 Monte Carlo samples & $S{=}4$ (RoboCloth area emitter, Bonn LLS),
                          $S{=}1$ (point lights) \\
Stage~2 representation  & Dense latent--geometry grid $2048^2\times46$
                          ($24$ latent $+\,16$ geometry $+\,6$ TBN-frame
                          channels), init $\mathcal{N}(0, 0.1^2)$ \\
Stage~2 geometry        & Per-pixel TBN (Gram--Schmidt) +
                          displacement MLP $[16,16,8]$ \\
Stage~2 regularization  & Progressive Gaussian blur,
                          $\sigma(t) = 2\cdot2^{-t/15000}$ \\
Stage~2 optimizer       & Adam (full precision),
                          lr${=}2\times10^{-3}$ (texture and geometry) \\
Stage~2 epochs          & $60$ (UBO2014 transfer) / $120$ (Bonn) /
                          $80$--$120$ (our captures, complexity-dependent) \\
Stage~2 loss            & Log-relative, $\rho_\text{ref}{=}0.05$ (UBO2014 transfer) \\
Per-channel scale       & $\boldsymbol{\beta}$, init $\mathbf{1}$ \\
PBR baseline            & Disney principled, $\mathbf{c}_b$+$\rho$+$m$+IOR+aniso+sheen+normal
                          at $2048^2$ \\
Test SPP                & $16$ \\
Metrics                 & PSNR, log-relative loss, qualitative renderings \\
\bottomrule
\end{tabular}%
}
\caption{Full implementation details. Settings are shared across
datasets unless noted.}
\label{tab:implementation_details}
\end{minipage}

\vspace{2em}
\begin{minipage}{\textwidth}
\centering
\resizebox{\textwidth}{!}{
\begin{tabular}{l l l}
\toprule
Path & Format & Contents \\
\midrule
\multicolumn{3}{l}{\textit{\texttt{materials/\{id\}/} --- one folder per material}} \\
\midrule
\texttt{hdr.tar}                      & TAR of 16-bit PNG & ${\sim}580$ cropped, polygon-masked linear-HDR frames \\
\texttt{observations\_structured.npz} & NumPy NPZ         & \texttt{rgbs} $(K,V,3)$ \texttt{uint16} radiance tensor; \texttt{xyz} $(V,3)$ \texttt{float32} \\
                                      &                   & point positions; \texttt{point\_ids} $(V,)$ \texttt{int32}; \texttt{cam\_pos}, \texttt{light\_pos} $(K,3)$ \\
\texttt{point\_positions.npz}         & NumPy NPZ         & \texttt{point\_ids} $(V,)$ and \texttt{positions} $(V,3)$ in base frame (m) \\
\texttt{rotated\_camera.json}         & JSON              & Per-frame camera-to-world pose: position and $3{\times}3$ rotation (turntable-aligned) \\
\texttt{scan\_log.json}               & JSON              & Per-scan acquisition log (camera and emitter ids, exposure) \\
\texttt{point\_metadata.json}         & JSON              & Surface-point count $V$ and per-material bookkeeping \\
\texttt{hdr\_crop\_bboxes.json}       & JSON              & Per-frame crop box used to cut each frame to the cloth \\
\texttt{bbox.json}                    & JSON              & Axis-aligned bounding box of the sample (base frame, m) \\
\texttt{unmatched\_scan\_ids.json}    & JSON              & Scan ids excluded from cross-view matching \\
\midrule
\multicolumn{3}{l}{\textit{\texttt{globals/} --- shared across all materials}} \\
\midrule
\texttt{emitter\_calibration.json}    & JSON & Directional emitter profile $\Lambda(\theta)$ \\
\texttt{camera\_factor.json}          & JSON & Radiometric scale $\kappa$ per exposure cohort \\
\texttt{sample\_size.json}            & JSON & Physical sample size $(w_m, l_m)$ per material id \\
\texttt{training\_list\_\{100,300,442,500\}.txt} & text & Training-split material ids at four corpus sizes \\
\texttt{test\_list\_\{100,300,442,500\}.txt}     & text & Held-out test material ids at four corpus sizes \\
\midrule
\multicolumn{3}{l}{\textit{Repository root}} \\
\midrule
\texttt{README.md}, \texttt{LICENSE}, \texttt{croissant.jsonld} & --- & Dataset card, CC BY 4.0 license, and Croissant (RAI) metadata \\
\bottomrule
\end{tabular}
}
\caption{Released Hugging Face dataset schema. \texttt{\{id\}} is the
integer material id; $K$ and $V$ are the per-material observation and
surface-point counts (\Cref{sec:obs_tensor}).}
\label{tab:schema}
\end{minipage}
\end{table*}

\section{Dataset Schema}
\label{supp:dataset_schema}

The dataset is released as a public Hugging Face dataset under the
CC BY 4.0 license. \Cref{tab:schema} documents its
repository layout: one folder per material under \texttt{materials/},
shared calibration and split files under \texttt{globals/}, and the
dataset card, license, and Croissant metadata at the repository root.
Floating-point quantities are in SI units (meters, radians) unless noted;
$K$ and $V$ denote the per-material observation and surface-point counts
introduced in \Cref{sec:obs_tensor}.

%% file: figures/datasize.tex
\begin{figure}[t]
  \centering
  \includegraphics[width=0.82\linewidth]{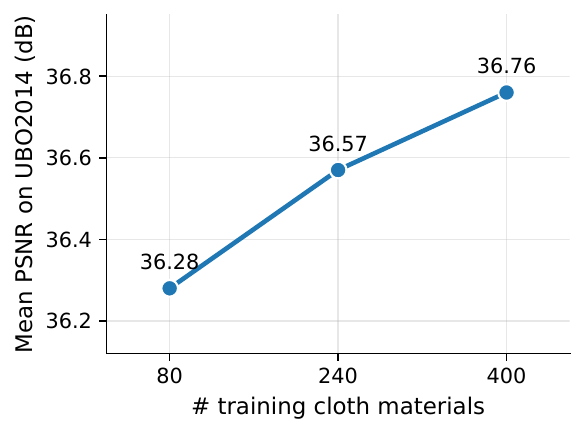}
  \caption{\textbf{Transfer PSNR vs.\ training-set size.} Mean
  UBO2014 transfer PSNR (over all $12$ transfer materials) against the
  number of training cloth materials ($80/240/400$, i.e.\ the
  $100/300/500$-material subsets). Accuracy rises consistently with more
  data and begins to saturate at our full corpus. The vertical axis is
  zoomed to make the trend visible.}
  \label{fig:datasize}
\end{figure}